\documentclass{article}
\usepackage{algorithm}
\usepackage[noend]{algpseudocode}
\usepackage{xcolor}
\usepackage{setspace}
\usepackage{algpseudocode}
\usepackage{amsmath}
\usepackage{subcaption}
\usepackage{amsmath}
\usepackage{amssymb}
\usepackage{caption}
\usepackage{algorithm}
\usepackage{algpseudocode}
\usepackage{amsmath}
\usepackage[export]{adjustbox}
\usepackage[utf8]{inputenc}
\usepackage{array}
\usepackage{color}
\usepackage{xcolor}
\usepackage{hyperref}
\usepackage{multirow}

\usepackage{arxiv}

\usepackage[utf8]{inputenc} 
\usepackage[T1]{fontenc}    
\usepackage{hyperref}       
\usepackage{url}            
\usepackage{booktabs}       
\usepackage{amsfonts}       
\usepackage{nicefrac}       
\usepackage{microtype}      
\usepackage{lipsum}
\usepackage{graphicx}
\graphicspath{ {./images/} }
\usepackage[numbers]{natbib}
\usepackage{algorithm}
\usepackage[noend]{algpseudocode}
\usepackage{xcolor}
\usepackage{setspace}
\usepackage{algorithm}
\usepackage{algpseudocode}
\usepackage{amsmath}
\usepackage{subcaption}

\usepackage{amsmath}
\usepackage{amssymb}
\usepackage{caption}
\usepackage{algorithm}
\usepackage{algpseudocode}
\usepackage{amsmath}
\usepackage[export]{adjustbox}
\usepackage[utf8]{inputenc}
\usepackage{array}
\usepackage{color}
\usepackage{amsthm}
\usepackage{hyperref}
\newtheorem{definition}{Definition}
\newtheorem{theorem}{Theorem}
\newtheorem{lemma}{Lemma}
\newtheorem{proofthm}{Proof}

\def\tsc#1{\csdef{#1}{\textsc{\lowercase{#1}}\xspace}}
\tsc{WGM}
\tsc{QE}
\tsc{EP}
\tsc{PMS}
\tsc{BEC}
\tsc{DE}

\title{Privacy-Preserving Federated Learning with Differentially Private Hyperdimensional Computing}

\author{
Fardin Jalil Piran \\
  School of Mechanical, Aerospace, and Manufacturing Engineering\\
  University of Connecticut\\
  Storrs, CT 06269 \\
  \texttt{fardin.jalil\_piran@uconn.edu} \\
   \And
 Zhiling Chen \\
  School of Mechanical, Aerospace, and Manufacturing Engineering\\
  University of Connecticut\\
  Storrs, CT 06269 \\
  \texttt{zhiling.chen@uconn.edu} \\
  \And
 Mohsen Imani \\
  Department of Computer Science\\
  University of California Irvine\\
  Irvine, CA 92697\\
  \texttt{m.imani@uci.edu} \\
  \And
Farhad Imani\\
  School of Mechanical, Aerospace, and Manufacturing Engineering\\
  University of Connecticut\\
  Storrs, CT 06269 \\
  \texttt{farhad.imani@uconn.edu} 
}

\begin{document}
\maketitle
\begin{abstract}
Federated Learning (FL) has become a key method for preserving data privacy in Internet of Things (IoT) environments, as it trains Machine Learning (ML) models locally while transmitting only model updates. Despite this design, FL remains susceptible to threats such as model inversion and membership inference attacks, which can reveal private training data. Differential Privacy (DP) techniques are often introduced to mitigate these risks, but simply injecting DP noise into black-box ML models can compromise accuracy, particularly in dynamic IoT contexts, where continuous, lifelong learning leads to excessive noise accumulation. To address this challenge, we propose Federated HyperDimensional computing with Privacy-preserving (FedHDPrivacy), an eXplainable Artificial Intelligence (XAI) framework that integrates neuro-symbolic computing and DP. Unlike conventional approaches, FedHDPrivacy actively monitors the cumulative noise across learning rounds and adds only the additional noise required to satisfy privacy constraints. In a real-world application for monitoring manufacturing machining processes, FedHDPrivacy maintains high performance while surpassing standard FL frameworks --- Federated Averaging (FedAvg), Federated Proximal (FedProx), Federated Normalized Averaging (FedNova), and Federated Optimization (FedOpt) --- by up to 37\%. Looking ahead, FedHDPrivacy offers a promising avenue for further enhancements, such as incorporating multimodal data fusion.
\end{abstract}


\textbf{DOI:} \href{https://doi.org/10.1016/j.compeleceng.2025.110261}{https://doi.org/10.1016/j.compeleceng.2025.110261}

\begin{figure}
    \centering
    \includegraphics[width=\textwidth]{ 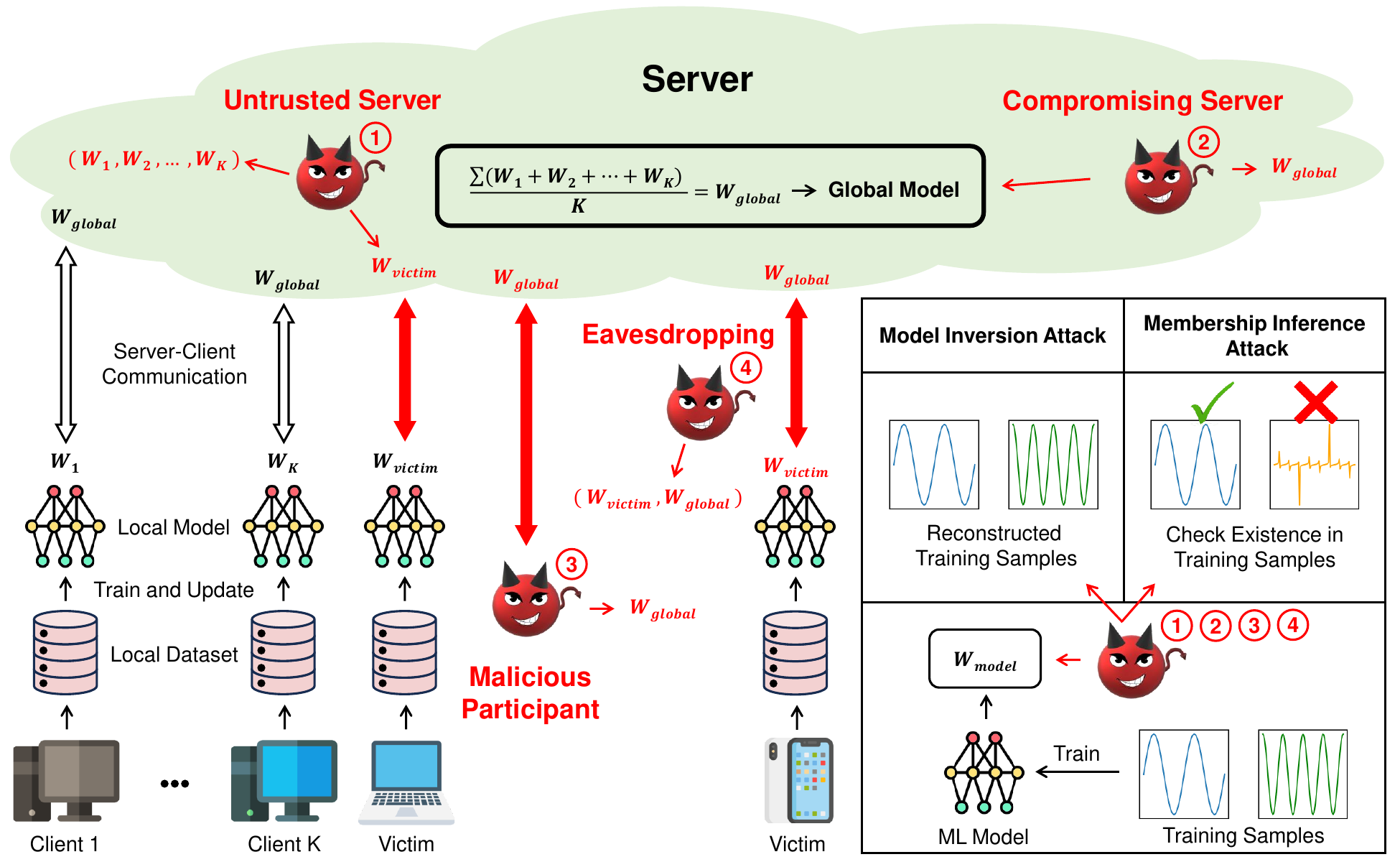}
        \caption{Federated learning structure with potential attacks. A federated learning setup where artificial intelligence and machine learning models are trained locally, with parameters aggregated on a central server. Potential attacks include eavesdropping, malicious participants, untrusted servers, and server breaches, all of which may expose local or global models. These attacks exploit model inversion and membership inference techniques to extract sensitive information about the training samples.}
    \label{fig:fl}
\end{figure}

 \begin{figure}
    \centering
    \includegraphics[width=\textwidth]{ 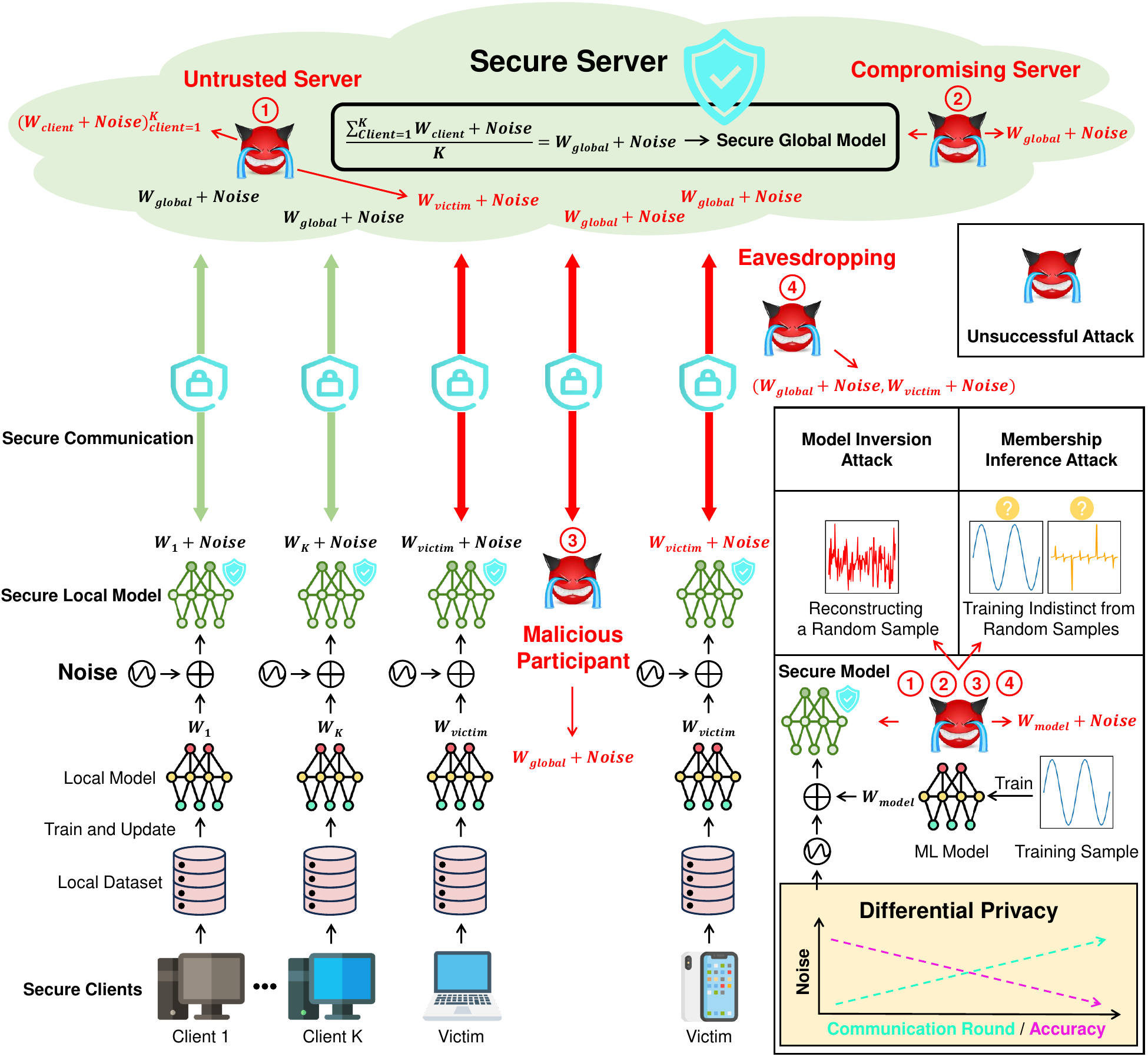}
        \caption{Secure federated learning framework with differential privacy and accuracy trade-off. A federated learning framework enhanced with differential privacy noise added to clients' models. This mechanism helps prevent adversaries from reconstructing training samples or distinguishing original data from random data, thereby safeguarding client information. However, the cumulative noise added over multiple training rounds introduces an accuracy trade-off, gradually impacting model performance.}
    \label{fig:dp}
\end{figure}

\section{Introduction}
\label{sec:Introduction}

The rapid expansion of the Internet of Things (IoT) has led to the widespread
integration of sensing and computing technologies, connecting a vast array of
devices to support applications in areas, including smart
cities~\cite{bhattacharya2022review} and digital
manufacturing~\cite{wang2020machine}. This interconnected landscape has
facilitated the emergence of intelligent IoT applications, where Artificial
Intelligence (AI) models play a crucial role in deriving actionable insights
from the data produced by IoT devices, such as in traffic
management~\cite{zeng2022federated}. Traditionally, these AI-driven tasks were
handled by centralized data centers~\cite{sun2019application}. However, this
approach is increasingly challenged by the impracticality of transmitting large
volumes of IoT data to distant servers and the heightened risk of privacy
breaches~\cite{nguyen2021federated}. Relying on third-party data centers for AI
processing can compromise sensitive information, including financial and
healthcare records~\cite{shirvani2023survey}. Therefore, there is a pressing
need for novel AI methods that not only protect privacy but also enhance the
efficiency and intelligence of IoT networks and applications.

Federated Learning (FL) has emerged as a powerful approach for building
intelligent, privacy-preserving systems within the IoT. FL operates as a
decentralized AI model to be trained directly on IoT devices, referred to as
clients, without the need to transfer raw data to a central
server~\cite{rodriguez2023survey}. As shown in Figure~\ref{fig:fl}, IoT devices act
as clients in this architecture, collaborating with a central server to refine
a global model. The process begins with the server initializing the global
model with set parameters, which are then distributed to the clients. Each
client uses its locally generated data to update the model and sends the
updates back to the server. The server aggregates the locally updated models,
improving the global model in an iterative manner. This decentralized approach
leverages the computational capabilities of distributed IoT devices, enhancing
training efficiency while ensuring data
privacy~\cite{mcmahan2017communication}.

In IoT environments, data is continuously generated and frequently change. This dynamic nature necessitates the implementation of continuous learning within the FL framework. Continuous learning, or lifelong learning, allows models to continuously incorporate new data, enabling them to adapt to evolving environments and remain effective in real-time applications. By regularly updating the model with new data streams, FL ensures that the global model stays current and responsive to changing conditions across distributed devices~\cite{le2021federated}. This capability is particularly valuable in IoT systems, where the model must continually learn from new information to maintain high performance in dynamic, real-world scenarios.

While FL improves privacy by keeping user data on individual devices rather than transmitting it to a central server, it does not fully eliminate privacy risks. If an attacker gains access to either the clients' or server's models, they can exploit these models to perform model inversion or membership inference attacks, extracting sensitive and confidential information about the training samples. In a model inversion attack, the attacker analyzes the model's outputs or gradients to reconstruct the original training data, effectively revealing sensitive information used to train the model~\cite{fredrikson2015model}. For example, by querying the trained model, an adversary can reverse-engineer the data to infer specific details about the training samples. In contrast, a membership inference attack allows an adversary to determine whether a particular data point was included in the training set~\cite{shokri2017membership}. This type of attack can reveal the presence or absence of specific individuals in the dataset, leading to severe privacy breaches. 

Both attacks pose significant risks to user privacy, even in seemingly secure
environments. Techniques such as generative regression neural networks have
shown that it is possible to extract sensitive information from shared model
parameters~\cite{ren2022grnn}. Even in black-box scenarios, where the internal
workings of the model are hidden, attackers can infer whether certain
individuals were part of the training set or recover features of their
data~\cite{fredrikson2015model}. As depicted in Figure~\ref{fig:fl}, these
vulnerabilities enable adversaries to exploit FL systems and compromise the
confidentiality of participants~\cite{guo2023fast}.

Attackers can access ML models through four primary methods. The first method
is eavesdropping, where adversaries intercept communication between clients and
the central server. As illustrated in Figure~\ref{fig:fl}, Since FL involves
multiple rounds of communication between clients and the server, unprotected
channels present a high risk of interception, potentially allowing attackers to
access the global and local models and their parameters~\cite{yuan2021beyond}.
The second method is by posing as a malicious participant, where attackers
disguise themselves as legitimate clients to interact directly with the central
server~\cite{fortino2020resiot}. This gives them access to the global model,
enabling them to extract sensitive information or infer the presence of
specific data. The third method is server compromise, where attackers
successfully hack into the central server, gaining control over the global
model and exposing sensitive data collected from multiple
clients~\cite{khan2018iot}. Lastly, in scenarios involving an untrusted server,
clients may fear that the server itself could analyze the local models it
receives, potentially identifying sensitive training data or determining if
specific data was used in training~\cite{omolara2022internet}. In all four
cases, attackers aim to exploit the models by using model inversion and
membership inference attacks to extract sensitive information about the
training samples, whether they gain access to the global or local models.

As vulnerabilities in ML models within FL frameworks are identified, the need
for robust defense mechanisms becomes increasingly critical. Recent studies
have investigated various strategies to enhance privacy in FL. One widely
adopted method is anonymization, which involves generalizing or redacting
specific attributes in a dataset to prevent the easy identification of
individual records. However, with the advent of sophisticated privacy attacks,
such as model inversion~\cite{fredrikson2015model}, which can reconstruct
training data even with limited access to the models, traditional anonymization
techniques have proven inadequate. Attackers often find ways to circumvent
these defenses, particularly in high-dimensional datasets where anonymization
fails to offer strong protection against the disclosure of sensitive
attributes.

Homomorphic Encryption (HE) has been introduced as a robust privacy-preserving
technique that allows computations to be performed directly on encrypted data,
eliminating the need for decryption~\cite{aono2017privacy}. While HE offers
strong theoretical privacy protection by enabling model training on encrypted
datasets, its practical application in modern ML systems is hindered by
significant computational overhead. This challenge is particularly evident in
deep learning scenarios, where the processing of large datasets and complex
models is common. Consequently, HE is often more applicable in situations where
models are pre-trained and deployed as services~\cite{zhu20182p}. Another
promising technique is Secure Multiparty Computation (SMC), which enables
multiple parties to jointly compute a function using their private inputs
without disclosing these inputs to one another or a central server. This method
removes the need for a trusted third party and provides strong privacy
guarantees in theory~\cite{sayyad2020privacy}. However, SMC also faces
challenges due to its high computational and communication demands, making it
less feasible to train intricate models in FL environments.

Differential Privacy (DP) has emerged as a practical
and effective solution for safeguarding privacy in AI systems, particularly
within FL frameworks. By introducing carefully calibrated noise to data or
models, DP ensures that individual records remain private during analysis,
mitigating risks from model inversion, and membership inference
attacks~\cite{cyffers2022privacy,kim2021federated}. This approach has been
successfully implemented by companies like Google, Microsoft, and Apple to
enhance user privacy while preserving data
utility~\cite{ding2019novel,erlingsson2014rappor,apple2017learning}. In FL
structures, as depicted in  Figure~\ref{fig:dp}, DP can be applied to local models
before transmission to the central server, ensuring that both local and global
models remain secure against potential breaches, even in scenarios involving
untrusted servers. This configuration also safeguards communication channels,
rendering eavesdropping and malicious participant attacks ineffective. Compared
to methods like anonymization, HE, and SMC, DP strikes an optimal balance
between privacy and practicality, making it a preferred choice for real-world
applications.

While DP is highly effective at safeguarding sensitive
information and maintaining data utility, it presents challenges in balancing
privacy and accuracy~\cite{fardin2024privacy}. In the context of continuous FL
combined with DP, noise is added during each communication round to protect
privacy, but this cumulative noise can significantly degrade model performance
over time. Excessive noise reduces accuracy, while insufficient noise increases
the risk of privacy breaches. To address this, it is crucial to develop
privacy-preserving mechanisms capable of tracking the noise accumulation across
communication rounds and adding only the necessary noise in each round to
guarantee privacy without unnecessarily compromising accuracy. Moreover,
evaluating privacy leakage for a given privacy budget is essential before
deploying models or releasing datasets~\cite{ren2022grnn}. This challenge is
further complicated by the black-box nature of Machine Learning (ML) models, where
their internal processes lack transparency~\cite{khalid2023privacy}. In
safety-critical applications, where both explainability and privacy are
critical, this lack of traceability becomes a significant issue. Therefore, the
development of adaptive noise management strategies within FL frameworks is
essential to maintain a balance between privacy and accuracy, ensuring the
effectiveness of privacy-preserving mechanisms without undermining model
performance~\cite{shaik2023remote}.

We propose Federated HyperDimensional computing with Privacy-preserving (FedHDPrivacy), an eXplainable Artificial Intelligence (XAI) framework for FL that addresses the privacy challenges in dynamic IoT environments. By combining DP and HyperDimensional computing (HD), FedHDPrivacy ensures robust privacy protection while maintaining high model accuracy during the aggregation process. The framework leverages XAI techniques to dynamically track the cumulative noise added in previous communication rounds and adjust the noise in subsequent rounds, introducing only the minimal necessary noise to guarantee privacy without compromising accuracy. This adaptive approach prevents excessive noise accumulation that could degrade model performance, securing both client and server models at every training round. Furthermore, FedHDPrivacy is designed to support continuous learning in IoT systems, enabling the global model to remain effective and responsive in dynamic environments. By safeguarding sensitive information and preserving model utility, FedHDPrivacy offers a secure, efficient, and explainable solution tailored to the real-world challenges of FL in IoT scenarios. Our contributions in this paper are summarized as follows:

\begin{enumerate}
    \item We introduce FedHDPrivacy, a novel and explainable framework that integrates HD with DP within a FL structure, ensuring robust privacy protection and model interpretability.
    \item Our framework accurately tracks cumulative noise from previous communication rounds and determines the incremental noise required, ensuring compliance with DP guarantees while avoiding unnecessary noise addition.
    \item FedHDPrivacy demonstrates adaptability to dynamic IoT environments by enabling continuous learning without performance degradation, effectively balancing privacy, accuracy, and efficiency.
    \item The proposed framework outperforms existing FL methods --- Federated Averaging (FedAvg), Federated Proximal (FedProx), Federated Normalized Averaging (FedNova), and Federated Optimization (FedOpt) --- in accuracy (up to 37\%) and computational efficiency, as validated through extensive experiments using real-world IoT manufacturing data.
    \item We provide a comprehensive analysis of hyperparameter effects, including hypervector size and privacy budget, on model performance, training time, and energy efficiency, offering insights into practical implementations of privacy-preserving FL systems.
\end{enumerate}

The remainder of this paper is structured as follows: Section~\ref{sec:Related Work} reviews existing FL frameworks, examines DP as a mechanism for safeguarding ML models in FL, and explores the integration of HD with FL and DP. In Section~\ref{sec:Preliminary}, we define key concepts related to HD and DP. Section~\ref{sec:Research Methodology} outlines the proposed FedHDPrivacy framework and its implementation. Section~\ref{sec:Experimental Design} details the experimental setup used in a real-world IoT scenario. The results of the experiments are presented and analyzed in Section~\ref{sec:Experimental Results}. Finally, Section~\ref{sec:Conclusions and Future Work} summarizes the contributions of this work in developing privacy-preserving FL models for IoT and offers suggestions for future research directions. The variables used throughout this work are summarized in Table~\ref{table:notation}.

\section{Related Work}
\label{sec:Related Work}
\subsection{Federated Learning}
\label{sec:RW Federated Learning}

FL offers substantial benefits for IoT applications by
enabling local data processing on devices while transmitting only model updates
to a central server. This decentralized approach minimizes the risk of exposing
sensitive information to third parties and ensures compliance with strict data
privacy regulations, such as the general data protection
regulation~\cite{goddard2017eu}. FL is particularly well-suited for secure IoT
systems, as it leverages the computational power and diverse datasets of IoT
devices to accelerate training and achieve higher accuracy compared to
centralized AI methods, which often struggle with limited data and
resources~\cite{imteaj2021survey}. Additionally, FL's decentralized structure
enhances scalability, making it ideal for expanding intelligent networks.

The evolution of FL has seen the development of
several frameworks addressing distributed optimization challenges and the
heterogeneity of client data and resources. Federated Stochastic Gradient
Descent (FedSGD) and FedAvg, introduced by McMahan
et al. ~\cite{mcmahan2017communication}, laid the foundation for FL. While FedSGD
aggregates gradients from multiple clients, FedAvg simplifies the process by
averaging model parameters across clients, offering an effective approach for
distributed learning. To address client heterogeneity and improve convergence,
Li et al. ~\cite{li2020federated} proposed FedProx, which
introduces a proximal term in the local objective functions. FedNova, developed by Wang
et al. ~\cite{wang2020federated}, resolves objective inconsistencies by
normalizing client updates, ensuring a more robust and consistent aggregation
process. Additionally, Federated Adam (FedAdam), introduced by Reddi
et al. ~\cite{reddi2020adaptive}, extends the Adam optimizer to the federated
setting by adapting learning rates for each client through second-order moment
estimates, improving convergence in non-identically distributed data scenarios.
Furthermore, the FedOpt framework, also proposed by
Reddi et al. ~\cite{reddi2020adaptive}, generalizes client and server
optimization strategies, allowing for the integration of different optimizers
such as SGD and Adam on the server side. FedOpt provides flexibility in tuning
the aggregation process, leading to better performance across diverse FL
scenarios.

These FL frameworks enable distributed learning among clients without transmitting raw data, thereby safeguarding confidential information. However, they remain vulnerable to model inversion and membership inference attacks, highlighting the need for DP to secure these frameworks against such threats.

\begin{table}[h!]
\centering
\caption{Notation and description of the variables used in this paper.
}
\renewcommand{\arraystretch}{1.4}
\begin{tabular}{c c}
\hline
Notation & Description   \\
\hline

$\mathbf{X}^{s}_{k,r}$ & Training data generated by client $k$ for class $s$ at round $r$\\
$\vec{H}^{s}_{k,r}$  & Hypervectors from client $k$ for class $s$ at round $r$  \\
$\vec{C}^{s}_{k,r}$ &  Class $s$ hypervector for client $k$ at round $r$\\
$\{{\vec{C}}^{s}_{g,r}\}_{s=1}^{S}$ &  Global model at round $r$\\
$\{\tilde{{\vec{C}}}^{s}_{k,r}\}_{s=1}^{S}$ &  Secure local model for client $k$ at round $r$\\
$\{\tilde{{\vec{C}}}^{s}_{g,r}\}_{s=1}^{S}$ &  Secure global model at round $r$\\
$\sigma_{dp}^{2}$ & Noise level (Variance)   \\
$\Delta$ & Sensitivity\\
$\epsilon$  & Privacy budget  \\
$\delta$ & Privacy loss threshold  \\

$\vec{\xi}_{k}^{r}$ & Required noise for client $k$'s local model at round $r$\\
$\vec{\Psi}_{k}^{r}$ & Cumulative noise in client $k$'s local model at round $r$\\
$\vec{\Gamma}_{k}^{r}$ & Additional noise added to client $k$'s local model at round $r$\\

$\vec{\xi}_{g}^{r}$ & Required noise for the global model at round $r$\\
$\vec{\Psi}_{g}^{r}$ & Cumulative noise in the global model at round $r$\\
$\vec{\Gamma}_{g}^{r}$ & Additional noise added to the global model at round $r$\\

 \hline
\end{tabular}
\label{table:notation}
\end{table}

\subsection{Differential Privacy}
\label{sec:RW Differential Privacy}

Several studies have investigated the integration of
DP within FL frameworks to enhance system security. Gong et al.  proposed a
framework that combines DP with HE to protect client gradients from the central
server, though challenges remain in fully preventing information leakage even
with DP~\cite{gong2020privacy}. Wu et al.  applied DP to multi-task learning in
FL by perturbing parameters with Gaussian noise to preserve privacy; however,
their method remains susceptible to attacks such as label-flipping and model
inversion~\cite{wu2020theoretical}. Cyffers et al.  introduced a decentralized FL
approach utilizing a relaxed form of local DP, enabling data analysis across
devices while balancing privacy and utility~\cite{cyffers2022privacy}.
Nevertheless, the decentralized nature of this protocol leaves it vulnerable to
data poisoning and other security threats~\cite{el2022differential}. Tran et al. 
developed a secure decentralized training framework that eliminates the need
for a central server by rotating the master node role among
clients~\cite{tran2021efficient}, but this approach struggles to fully defend
against collusion attacks~\cite{el2022differential}. Zhao et al.  proposed a
method involving the sharing of partial gradient parameters combined with
Gaussian noise and introduced an additional proxy to enhance client
anonymity~\cite{zhao2021anonymous}. While this technique strengthens privacy,
it requires further evaluation to assess its resilience against inference
attacks. Yin et al.  suggested a privacy-preserving strategy using functional
encryption and Bayesian DP to secure client--server communication and monitor
privacy leakage~\cite{yin2021privacy}. However, this method relies on a trusted
third party, which introduces vulnerabilities to model inversion attacks. Other
works have focused on adaptive encoding methods to dynamically adjust noise
addition based on input data distribution, minimizing utility loss and
achieving improved accuracy~\cite{gao2024amoue}. DP techniques have also been
proposed for preserving information in cyber--physical systems by introducing
random noise to generate anonymity, though these methods can impact data
quality~\cite{basak2023dppt}. Approaches leveraging three-way attribute
decisions in IoT systems have shown potential in stabilizing datasets and
improving privacy by categorizing sensitive attributes before applying
DP~\cite{ali2022privacy}. Recent advancements have further integrated dynamic
privacy budget adjustment and adaptive weighting mechanisms to address
challenges posed by heterogeneous IoT data, achieving improved privacy
preservation and performance~\cite{zheng2025awe}.

Given the dynamic nature of IoT systems, FL must support continuous training over time, adapting to newly generated data in each communication round. While DP provides privacy by adding noise in every round, the cumulative effect of this noise can degrade model accuracy over time. To address this challenge, an XAI model is needed to monitor and manage the cumulative noise from previous rounds. By accurately tracking this noise, the XAI model can calculate the additional noise required for privacy as the difference between the target noise level and the cumulative noise already introduced. This adaptive approach helps maintain an optimal balance between privacy and accuracy throughout the FL process, ensuring the framework remains effective in dynamic, real-world IoT environments.

\subsection{Hyperdimensional Computing}
\label{sec:RW Hyperdimensional computing}

HD draws significant inspiration from the human brain,
leveraging high-dimensional vectors to simulate cognitive
functions~\cite{imani2023hierarchical,poduval2022graphd,zou2022memory}. This
advanced approach enables HD to mimic brain-like methods for processing,
analyzing, and storing information across diverse cognitive
tasks~\cite{JalilPiran2023,chen2023brain,zou2021scalable}. Several studies have
explored the integration of HD within FL frameworks. Zhao
et al. ~\cite{zhao2022fedhd} proposed an FL framework utilizing HD, demonstrating
the model's energy and communication efficiency. While their work highlighted
HD's robustness against noise, it did not address the privacy implications. Li
et al. ~\cite{li2024hyperfeel} introduced an HD-based FL framework designed to
reduce computational and communication overhead during local training and model
updates. Hsieh et al. ~\cite{hsieh2021fl} bipolarized model parameters in an
HD-based FL framework to minimize communication costs, yet their work did not
incorporate specific defenses against model inversion and membership inference
attacks. As an XAI model, HD is inherently susceptible to model inversion and
inference attacks~\cite{piran2024explainable}, emphasizing the need for
privacy-preserving mechanisms to protect HD-based FL systems from adversarial
threats.

 Some studies have explored privacy-preserving
techniques specifically for HD models. Hernandez-Cano
et al. ~\cite{hernandez2021prid} implemented intelligent noise injection by
randomizing insignificant features in the original feature space. However,
their approach does not clearly define the required noise level to achieve a
specific privacy guarantee. Khaleghi et al. ~\cite{khaleghi2020prive} introduced
DP into HD models to enhance security but did not predict the impact of noise
on model accuracy. Piran et al. ~\cite{piran2024explainable} examined the
transparency of combining DP with HD models and evaluated noise's effect on HD
performance at various privacy levels, though their study did not address these
considerations within an FL framework. Consequently, there is a critical need
to explore the transparency and explainability of HD models within the context
of FL. This includes developing techniques for precisely adding noise to both
client models and the global model, ensuring robust data security while
minimizing the impact on model accuracy.

\section{Preliminary}
\label{sec:Preliminary}
 \begin{figure}
    \centering
    \includegraphics[width=\textwidth]{ 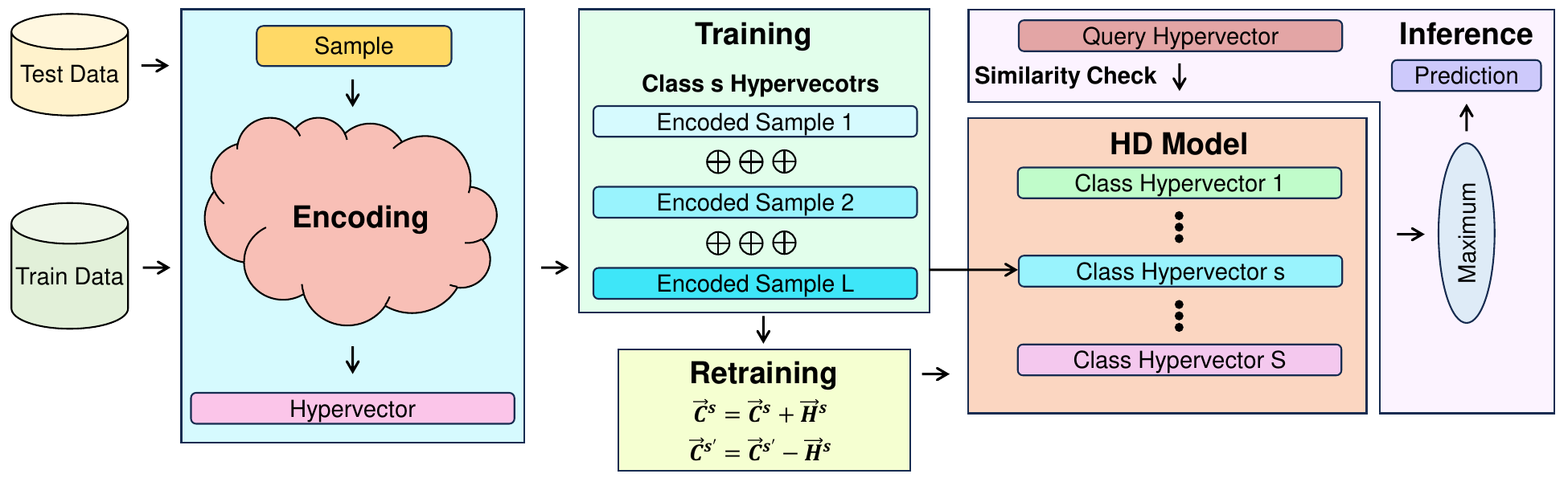}
        \caption{\textcolor{black}{Overview of the hyperdimensional computing framework. This framework illustrates the hyperdimensional computing process, covering encoding, training, inference, and retraining phases. In encoding, raw data are transformed into hypervectors. During training, hypervectors from the same class are aggregated to create class hypervectors. Inference involves comparing query hypervectors to class hypervectors for similarity, while retraining adjusts class hypervectors in response to misclassifications, improving model accuracy.}}
    \label{fig:hd}
\end{figure}

\subsection{Hyperdimensional Computing}
\label{sec:P Hyperdimensional Computing}

HD is a computational paradigm that leverages high-dimensional representations
to perform complex information processing
tasks~\cite{hoang2024hierarchical,rescsanski2023anomaly}. By employing
large-scale vectors, HD captures intricate patterns and relationships in data,
making it particularly well-suited for modeling and analyzing cognitive
processes~\cite{chen2021joint,poduval2021robust}. The HD framework, illustrated
in  Figure~\ref{fig:hd}, is structured into four key phases: encoding, training,
inference, and retraining. Each phase is meticulously designed to model
different aspects of cognitive processing. The process begins with the encoding
phase, where input data are transformed into high-dimensional hypervectors,
forming the foundation for the subsequent training phase. During training,
hypervectors are aggregated to create distinct class representations, crucial
for the model's ability to identify and classify new data during the inference
phase. The cycle concludes with the retraining phase, which allows for
continuous refinement and improvement of the model based on new data and
insights. This dynamic process mirrors the brain's capacity for learning and
adaptation, showcasing HD's ability to handle complex learning tasks with
remarkable accuracy and flexibility.

\subsubsection{Encoding}
\label{sec:P HD Encoding}
The encoding phase is essential for transforming input data into hypervectors,
which form the foundation of the high-dimensional computational framework. This
process ensures that every component of a hypervector plays an equal role in
representing the encoded information, thereby maintaining a balanced and
comprehensive data representation. An input sample, denoted as $\mathbf{x}$, is
mapped into a hypervector $\vec{H}$ as shown in~\eqref{eq:encoding}. This
mapping follows the principles of the random vector functional link model, a
technique that generates random projections to map input features into a
higher-dimensional space~\cite{kleyko2020density,JMLR:v24:23-0300}. This model
enhances the representational capacity by introducing non-linearities, ensuring
that the transformed hypervectors effectively capture the complexity of the
input data. After this transformation, the elements of the hypervector are
binarized.
\begin{equation} 
\phi : \mathbf{x} \rightarrow \vec{H}
\label{eq:encoding}
\end{equation}
\subsubsection{Training}
\label{sec:P HD Training}

In the training phase, class hypervectors, denoted as $\vec{C}^s$, are created for each class $s$, where $s$ ranges from $1$ to $S$ (the total number of classes). These hypervectors are formed by summing all associated hypervectors for each class, as described in~\eqref{eq:formClassHypervector}. Here, $\vec{H}^s$ represents the hypervectors linked to class $s$. This summation combines the features of all training samples within a class into a single high-dimensional representation, crucial for accurate classification. The aggregation process captures the shared characteristics of each class, enabling the HD model to effectively distinguish between different classes during inference. Creating class hypervectors is a key step in the HD training process, laying the groundwork for accurate pattern recognition and prediction capabilities.
\begin{equation} 
\vec{C}^s = \sum \vec{H}^s
\label{eq:formClassHypervector}
\end{equation}

\subsubsection{Inference}
\label{sec:P HD Inference}

In the inference phase, the class of a query hypervector is identified by comparing it to the class hypervectors generated during training. This comparison relies on cosine similarity, a measure that determines how closely two vectors are aligned in the high-dimensional space. Cosine similarity is calculated by taking the dot product of the two vectors and normalizing it by the product of their magnitudes. During this phase, the HD model computes the similarity between the query hypervector $\Vec{H}^{q}$ and each class hypervector $\Vec{C}^{s}$. The class associated with the hypervector that shows the highest similarity to $\Vec{H}^{q}$ is selected as the most likely classification for the query. This method allows the HD model to classify new data by applying the patterns learned during training and exploiting the spatial relationships inherent in high-dimensional vectors.
\begin{equation}
\text{Cos}(\Vec{C}^{s}, \Vec{H}^{q}) = \frac{\Vec{C}^{s} \cdot \Vec{H}^{q}}{||\Vec{C}^{s}|| \cdot ||\Vec{H}^{q}||}
\label{eq:inference similarity}
\end{equation}
\subsubsection{Retraining}
\label{sec:P HD Retraining}

Retraining plays a crucial role in enhancing the accuracy and adaptability of the HD model, particularly in continuous learning environments like FL. Instead of retraining the model from scratch with all previous data, retraining in HD focuses on correcting misclassifications by fine-tuning the model using new training samples. This process involves comparing the hypervectors of new data points with the existing class hypervectors. If a hypervector, $\Vec{H}^{s}$, is mistakenly classified into an incorrect class $s^{\prime}$ rather than its true class $s$, the model is updated to correct this error. The class hypervectors are adjusted by adding the hypervector to its correct class and subtracting it from the incorrect one, as expressed in the following equations:
\begin{equation}
\begin{aligned}
\Vec{C}^s  &\leftarrow \Vec{C}^s + \Vec{H}^{s} \\
\Vec{C}^{s^{'}} &\leftarrow \Vec{C}^{s^{'}} - \Vec{H}^{s}
\end{aligned}
\end{equation}

This adjustment refines the class hypervectors by integrating the correct information into the appropriate class while removing the influence from the incorrect classification. By continuously applying these updates, the HD model incrementally improves its classification accuracy, becoming better suited to handle changes in data distribution or the introduction of new patterns. This iterative process of refinement is particularly valuable in FL, where it enables the model to adapt to new data without the need to start the training process anew, thereby making HD a highly effective option for continuous learning.

\subsection{Differential Privacy}
\label{sec:P Differential Privacy}

In this work, we address the practical implementation of DP within FL frameworks, specifically tailored for IoT environments. To ensure robust privacy protection, we utilize DP mechanisms to introduce noise in a controlled manner, governed by key parameters such as the privacy budget ($\epsilon$) and the privacy loss threshold ($\delta$). The following definitions and formulations provide the foundational understanding of DP concepts and their application in this work.

\textcolor{black}{
\begin{definition}\label{def:DP}
A random algorithm \(M\) satisfies \((\epsilon, \delta)\)-DP if, for any two datasets \(I_1\) and \(I_2\) differing by only a single data point, the following condition holds:
\begin{equation}
\mathbb{P}[M(I_1)] \le exp(\epsilon) \cdot \mathbb{P}[M(I_2)] + \delta 
\end{equation}
This ensures that the presence or absence of a single data point in the dataset has a minimal effect on the algorithm’s output, thereby safeguarding individual privacy.
\end{definition}
}
\textcolor{black}{
\begin{definition}\label{def:Gaussian Mechanism}
The Gaussian Mechanism applies to any function \(f: I \to \mathbb{R}^{D}\) by adding Gaussian noise with mean 0 and variance \(\sigma_{dp}^2\) to its output. This noise obscures the influence of any single data point, ensuring compliance with DP:
\begin{equation}
M(I) = f(I) + \mathcal{N}(0,\sigma_{dp}^2)
\label{eq:eq_dp}
\end{equation}
\end{definition}
}
\textcolor{black}{
\begin{definition}\label{def:Sensitivity}
The sensitivity of a function \(f\), denoted as \(\Delta f\), is the maximum difference in \(f\)'s output when applied to any two datasets \(I_1\) and \(I_2\) that differ by a single element:
\begin{equation}
\Delta f = \max_{I_1, I_2} \|f(I_1) - f(I_2)\|
\label{eq:sensitive}
\end{equation}
This metric quantifies the impact a single data point can have on the function’s output.
\end{definition}
}
\textcolor{black}{
\begin{theorem}\label{thm:theorem1}
For a mechanism to satisfy \((\epsilon, \delta)\)-DP, the variance of the added Gaussian noise, \(\sigma_{dp}^2\), must satisfy:
\begin{equation}
\sigma_{dp}^2 > 2 \ln \frac{1.25}{\delta} \cdot \frac{\Delta f^2}{\epsilon^2}
\label{eq:sigma_dp}
\end{equation}
where \(\Delta f\) represents the sensitivity of the function.
\end{theorem}
}

\textcolor{black}{These formulations provide the theoretical foundation for the DP mechanisms used in this work. Specifically, the Gaussian Mechanism and sensitivity calculations allow for precise control of noise introduction, ensuring privacy while minimizing its impact on model performance. By adhering to these principles, the proposed framework effectively balances privacy and utility, addressing the unique challenges of continuous learning in IoT-driven FL systems.}

\begin{algorithm}
\caption{Federated hyperdimensional computing algorithm.}
\label{alg:fedhd}

\begin{algorithmic}[1]
\Statex \hspace{0em} \textbf{Input:}  $\{\{\{\mathbf{X}^{s}_{k,r}\}_{s=1}^{S}\}_{r=1}^{R}\}_{k=1}^{K}$ \Comment{Training samples generated by $K$ clients at $R$ rounds for $S$ classes}
\Statex \hspace{0em} \textbf{Output:}  $\{{\vec{C}}_{g,R}^{s}\}_{s=1}^{S}$ \Comment{Global model after $R$ rounds training communication}
\Statex \hspace{0em} \textbf{Function Encoding:}
\begin{algorithmic}[1]
\State \hspace{0em} $\vec{H}^{s}_{k,r} \leftarrow sign(\phi(\mathbf{X}^{s}_{k,r}))$
\Comment{Transform input data into hypervectors}
\State \hspace{0em} \textbf{Return} $\{\vec{H}^{s}_{k,r}\}_{s=1}^{S}$
\end{algorithmic}
\Statex \hspace{0em}  \textbf{Function Training:}
\begin{algorithmic}[1]
\State \hspace{0em} ${\vec{C}}^{s}_{k,r} \leftarrow \sum \vec{H}^{s}_{k,r}$ \Comment{Aggregate hypervectors to form class hypervectors}
\State \hspace{0em} \textbf{Return} $\{{\vec{C}}^{s}_{k,r}\}_{s=1}^{S}$
\end{algorithmic}
\Statex \hspace{0em}  \textbf{Function Retraining:}
\begin{algorithmic}[1]
\State \hspace{0em} \textbf{if} Any misprediction 
\State \hspace{1.5em} $\Vec{C}^{s}_{k,r}  \leftarrow \Vec{C}^{s}_{k,r} + \Vec{H}^{s}_{k,r}$  \Comment{Correct  mispredictions}
\State \hspace{1.5em} $\Vec{C}^{s^{'}}_{k,r} \leftarrow \Vec{C}^{s^{'}}_{k,r} - \Vec{H}^{s}_{k,r}$
\State \hspace{0em} \textbf{end if}
\State \hspace{0em} \textbf{Return} $\{{\vec{C}}^{s}_{k,r}\}_{s=1}^{S}$
\end{algorithmic}
\Statex \hspace{0em}  \textbf{Function Aggregation:}
\begin{algorithmic}[1]
\State \hspace{0em} ${\vec{C}}^{s}_{g,r} \leftarrow  \frac{\sum_{k=1}^{K} \Vec{C}^{s}_{k,r}}{K} $  \Comment{Aggregate local models  to form the global model}
\State \hspace{0em} \textbf{Return} $\{{\vec{C}}^{s}_{g,r}\}_{s=1}^{S}$
\end{algorithmic}
\Statex \hspace{0em}  \textbf{Function First Round:}
\begin{algorithmic}[1]
\State \hspace{0em} \textbf{for} $k \in \{1, 2, \ldots, K\}$ \textbf{do in Parallel}
\State \hspace{1.5em} $\{\vec{H}^{s}_{k,1}\}_{s=1}^{S} \leftarrow$ \textbf{Encoding(}$\{\mathbf{X}^{s}_{k,1}\}_{s=1}^{S}$\textbf{)}  
\State \hspace{1.5em} $\{{\vec{C}}^{s}_{k,1}\}_{s=1}^{S} \leftarrow$ \textbf{Training(}$\{\vec{H}^{s}_{k,1}\}_{s=1}^{S}${)}  \Comment{Client k's local model}   
\State \hspace{0em} \textbf{Return} $\{\{{\vec{C}}^{s}_{k,1}\}_{s=1}^{S}\}_{k=1}^{K}$
\end{algorithmic}
\Statex \hspace{0em}  \textbf{Function Local Updating:}
\begin{algorithmic}[1]
\State \hspace{0em} \textbf{for} $k \in \{1, 2, \ldots, K\}$ \textbf{do in Parallel}
\State \hspace{1.5em} $\{\vec{H}^{s}_{k,r}\}_{s=1}^{S} \leftarrow$ \textbf{Encoding(}$\{\mathbf{X}^{s}_{k,r}\}_{s=1}^{S}$\textbf{)}  \Comment{Encode new generated training samples}
\State \hspace{1.5em} $\{{\vec{C}}^{s}_{k,r}\}_{s=1}^{S} \leftarrow$ \textbf{Retraining(}$\{{\vec{C}}^{s}_{g,r-1}\}_{s=1}^{S},\{\vec{H}^{s}_{k,r}\}_{s=1}^{S}${)}\Comment{Update local models by retraining the global model}
\State \hspace{0em} \textbf{Return} $\{\{{\vec{C}}^{s}_{k,r}\}_{s=1}^{S}\}_{k=1}^{K}$
\end{algorithmic}
\Statex \hspace{0em} \textbf{Main Federated Hyperdimensional Computing:}
\begin{algorithmic}[1]
\State \hspace{0em} $\{\{{\vec{C}}^{s}_{k,1}\}_{s=1}^{S}\}_{k=1}^{K} \leftarrow$ \textbf{First Round(}$\{\{\mathbf{X}^{s}_{k,1}\}_{s=1}^{S}\}_{k=1}^{K}${)}  \Comment{Generate local models in the first round}
\State \hspace{0em} $\{{\vec{C}}^{s}_{g,1}\}_{s=1}^{S} \leftarrow$ \textbf{Aggregation(}$\{\{{\vec{C}}^{s}_{k,1}\}_{s=1}^{S}\}_{k=1}^{K}${)}  \Comment{Form the global model by aggregating the local models}
\State \hspace{0em} \textbf{for} $r \in \{2, \ldots, R\}$ \textbf{do}
\State \hspace{1.5em} $\{\{{\vec{C}}^{s}_{k,r}\}_{s=1}^{S}\}_{k=1}^{K} \leftarrow$ \textbf{Local Updating(}$\{{\vec{C}}^{s}_{g,r-1}\}_{s=1}^{S},\{\{\mathbf{X}^{s}_{k,r}\}_{s=1}^{S}\}_{k=1}^{K}${)}  \Comment{Update local models}
\State \hspace{1.5em} $\{{\vec{C}}^{s}_{g,r}\}_{s=1}^{S} \leftarrow$ \textbf{Aggregation(}$\{\{{\vec{C}}^{s}_{k,r}\}_{s=1}^{S}\}_{k=1}^{K}${)}  \Comment{Update the global model}
\end{algorithmic}
\end{algorithmic}
\end{algorithm}

\section{Research Methodology}
\label{sec:Research Methodology}

\textcolor{black}{In this study, we propose the FedHDPrivacy framework, which integrates FL, HD, and DP to securely facilitate collaborative learning in IoT environments. The framework leverages HD as an explainable model to analyze training data and classify new samples. The learning process is distributed across multiple clients, where each client trains an HD model locally and sends its updates to a central server. The server aggregates these local models to form a global model, which combines the knowledge of all clients and improves predictive accuracy.}

\textcolor{black}{To address privacy vulnerabilities inherent in FL, such as model inversion and membership inference attacks, DP is incorporated into the framework. Specifically, noise is added to local models before transmission, ensuring that individual training samples cannot be inferred from the updates. The global model, as an aggregation of these noisy local models, does not expose sensitive information, even in the event of server compromise or malicious participant activity. This strategy also protects against eavesdropping during communication and mitigates risks associated with untrusted servers.}

\textcolor{black}{The methodology consists of two main components. First, HD models are used to distribute the learning process across clients, with local training updates aggregated to build the global model. Second, an adaptive DP mechanism is employed to manage noise levels. The explainability of HD allows for precise determination of the noise required in each communication round. By calculating the cumulative noise introduced in previous rounds, the framework adjusts the noise added in subsequent rounds to meet privacy guarantees while minimizing the impact on model performance. This approach ensures that the global model achieves a balance between privacy protection and accuracy, avoiding unnecessary degradation caused by excessive noise.}

\begin{algorithm}
\caption{Federated hyperdimensional computing with privacy-preserving.}
\label{alg:fedprivacy}

\begin{algorithmic}[1]
\Statex \hspace{0em} \textbf{Input:}  $\{\{\{\mathbf{X}^{s}_{k,r}\}_{s=1}^{S}\}_{r=1}^{R}\}_{k=1}^{K}$ \Comment{Training samples for \(K\) clients over \(R\) rounds for \(S\) distinct classes}
\Statex \hspace{0em} \textbf{Output:}  $\{\tilde{{\vec{C}}}_{g,R}^{s}\}_{s=1}^{S}$ \Comment{Secure global model following \(R\) rounds of training communication}
\Statex \hspace{0em} \textbf{Function Required Noise:}
\begin{algorithmic}[1]
\State \hspace{0em} \textbf{if} First round
\State \hspace{1.5em} $\vec{\xi}_{k}^{r} \leftarrow \mathcal{N}\left(0, \frac {2D}{{\epsilon^2}} \ln{[1.25L]}\right)$
\State \hspace{0em} \textbf{Else} 
\State \hspace{1.5em} $\vec{\xi}_{k}^{r} \leftarrow \mathcal{N}\left(0, \frac{2D}{{\epsilon^2}} \ln{[1.25(r-1)KL + 1.25L]}\right)$
\State \hspace{0em} \textbf{Return} $\vec{\xi}_{k}^{r}$ \Comment{Required noise at round $r$ for clients}
\end{algorithmic}
\Statex \hspace{0em} \textbf{Function Cumulative Noise:}
\begin{algorithmic}[1]
\State \hspace{0em} \textbf{if} First round
\State \hspace{1.5em} $\vec{\Psi}_{k}^{r-1} \leftarrow 0 $
\State \hspace{0em} \textbf{Else} 
\State \hspace{1.5em} $\vec{\Psi}_{k}^{r-1} \leftarrow \mathcal{N}\left(0, \frac{2D}{{K\epsilon^2}} \ln{[1.25(r-2)KL + 1.25L]}\right)$
\State \hspace{0em} \textbf{Return} $\vec{\Psi}_{k}^{r-1}$ \Comment{Cumulative noise in local models at round $r-1$}
\end{algorithmic}
\Statex \hspace{0em} \textbf{Function Client Model:}
\begin{algorithmic}[1]
\State \hspace{0em} \textbf{for} $k \in \{1, 2, \ldots, K\}$ \textbf{do in Parallel}
\State \hspace{1.5em} \textbf{if} First round
\State \hspace{3em} $\{{\vec{C}}^{s}_{k,r}\}_{s=1}^{S}\ \leftarrow$  \textbf{Training(}\textbf{Encoding(}$\{\mathbf{X}^{s}_{k,r}\}_{s=1}^{S}$\textbf{))}
\State \hspace{1.5em} \textbf{Else}
\State \hspace{3em} $\{{\vec{C}}^{s}_{k,r}\}_{s=1}^{S}\ \leftarrow$ \textbf{Retraining(}\textbf{Encoding(}$\{\mathbf{X}^{s}_{k,r}\}_{s=1}^{S}$\textbf{))}
\State \hspace{0em} \textbf{Return} $\{\{{\vec{C}}^{s}_{k,r}\}_{s=1}^{S}\}_{k=1}^{K}\ $
\end{algorithmic}
\Statex \hspace{0em} \textbf{Function Secure Client:}
\begin{algorithmic}[1]
\State \hspace{0em} \textbf{for} $k \in \{1, 2, \ldots, K\}$ \textbf{do in Parallel}
\State \hspace{1.5em} \textbf{for} $s \in \{1, 2, \ldots, S\}$ \textbf{do in Parallel}
\State \hspace{3em} $\tilde{{\vec{C}}}^{s}_{k,r} \leftarrow  {\vec{C}}^{s}_{k,r} + \vec{\Gamma}_{k}^{r}$ 
\State \hspace{0em} \textbf{Return} $\{\{\tilde{{\vec{C}}}^{s}_{k,r}\}_{s=1}^{S}\}_{k=1}^{K}\ $
\end{algorithmic}
\Statex \hspace{0em} \textbf{Main FedHDPrivacy:}
\begin{algorithmic}[1]
\State \hspace{0em} \textbf{for} $r \in \{1, 2, \ldots, R\}$ \textbf{do}
\State \hspace{1.5em} $\{\{{\vec{C}}^{s}_{k,r}\}_{s=1}^{S}\}_{k=1}^{K} \leftarrow$ \textbf{Client Model(}$\{\{\mathbf{X}^{s}_{k,r}\}_{s=1}^{S}\}_{k=1}^{K}${)}  \Comment{Local models}
\State \hspace{1.5em} $\vec{\xi}_{k}^{r} \leftarrow$ \textbf{Required Noise(}$\epsilon${)}
\State \hspace{1.5em} $\vec{\Psi}_{k}^{r-1} \leftarrow$ \textbf{Cumulative Noise(}$\epsilon${)}
\State \hspace{1.5em} $\vec{\Gamma}_{k}^{r} \leftarrow \vec{\xi}_{k}^{r} - \vec{\Psi}_{k}^{r-1} $ \Comment{Additional noise added}
\State \hspace{1.5em} $\{\{\tilde{{\vec{C}}}^{s}_{k,r}\}_{s=1}^{S}\}_{k=1}^{K} \leftarrow $   \textbf{Secure Client(}$ \{\{{\vec{C}}^{s}_{k,r}\}_{s=1}^{S}\}_{k=1}^{K},\vec{\Gamma}_{k}^{r}${)} \Comment{Secure local models}
\State \hspace{1.5em} $\{\tilde{{\vec{C}}}^{s}_{g,r}\}_{s=1}^{S} \leftarrow$ \textbf{Aggregation(}$\{\{\tilde{{\vec{C}}}^{s}_{k,r}\}_{s=1}^{S}\}_{k=1}^{K}${)}  \Comment{Secure global model}

\end{algorithmic}
\end{algorithmic}
\end{algorithm}

\subsection{Federated Hyperdimensional Computing}
\label{sec:Federated Hyperdimensional Computing}

The HD model is particularly well-suited for FL due to its retraining phase, which enables the model to incorporate newly generated training samples in each round without forgetting the information from previous rounds. This continuous learning capability allows each client to update its HD model by retraining with new samples, ensuring that both new and previously learned information is retained. As a result, there is no need to retrain the HD model from scratch in every round, making it ideal for continuous learning scenarios within an FL framework.

We consider a scenario where the FL system operates in a continuous learning mode, with clients and the server communicating over an unlimited number of rounds, as illustrated in Algorithm~\ref{alg:fedhd}. In each round, every client uses a fixed number of samples, \(L\), to retrain their local model. Both local and global models consist of \(S\) class hypervectors, each representing a specific class. During the first round, each client builds its local model using \(L\) samples and then sends these local models to the server. The server aggregates the local models by calculating the element-wise average of the class hypervectors, as shown in Equation~\eqref{eq:aggregates} for $r=1$:
\begin{equation}
    \forall s \in \{1, 2, \dots, S\}: \quad {\vec{C}}_{g,r}^{s} = \frac{\sum_{k=1}^{K} {\vec{C}}_{k,r}^{s}}{K}
    \label{eq:aggregates}
\end{equation}

In subsequent rounds, each client retrains the global model using newly generated data. After receiving updates from the clients, the server aggregates the local models based on Equation~\eqref{eq:aggregates} and updates the global model. This process ensures that information from all training samples is preserved because the global model is continuously updated rather than being retrained from scratch. This approach not only aligns the global model closely with new samples, improving accuracy, but also retains the knowledge from previous rounds, thereby enhancing the overall robustness of the model.

\subsection{FedHDPrivacy}
\label{sec:FedHDPrivacy}

\textcolor{black}{The FedHDPrivacy framework ensures privacy in FL by incorporating DP into the exchange of class hypervectors between clients and the central server. Without a privacy-preserving mechanism, this shared information is vulnerable to model inversion and membership inference attacks, which could reveal sensitive training data. To mitigate these risks, Gaussian noise is added to the HD models, with the variance of the noise calculated based on key privacy parameters: the privacy budget (\(\epsilon\)), the privacy loss threshold (\(\delta\)), and the sensitivity of the function (\(\Delta f\)). As described in Section~\ref{sec:P Differential Privacy}, the noise variance is determined using Equation~\eqref{eq:sigma_dp}. The privacy loss threshold (\(\delta\)) is set as the inverse of the number of training samples, following the approach by Saifullah et al.~\cite{saifullah2024privacy}, while the sensitivity (\(\Delta f\)) is application-specific, and the privacy budget (\(\epsilon\)) is chosen based on the desired level of privacy.}

\textcolor{black}{Next, we address the noise required to secure both the local model (\(\vec{\xi}_{k}^{r}\)) for each client \(k\), and the global model (\(\vec{\xi}_{g}^{r}\)), in each round \(r\). Additionally, we account for the cumulative noise introduced in previous rounds, denoted as \(\vec{\Psi}_{k}^{r-1}\) for the local models and \(\vec{\Psi}_{g}^{r-1}\) for the global model. Finally, we compute the adjusted noise that needs to be added to maintain privacy. For the local models, this is calculated as \(\vec{\Gamma}_{k}^{r} = \vec{\xi}_{k}^{r} - \vec{\Psi}_{k}^{r-1}\), and for the global model, it is \(\vec{\Gamma}_{g}^{r} = \vec{\xi}_{g}^{r} - \vec{\Psi}_{g}^{r-1}\). This ensures that the privacy of all training samples is protected throughout the learning process. Additionally, noise is added to both the local and global models, as demonstrated in Algorithm~\ref{alg:fedprivacy}, to secure the FL structure and counteract model inversion and membership inference attacks.}

\textcolor{black}{In the first round of FL, clients independently train their local HD models using \(L\) training samples and share their noisy class hypervectors with the server. Theorem~\ref{thm:theorem2} defines the amount of noise required to ensure privacy while maintaining accuracy, with Proof~\ref{proof:proof1} deriving the sensitivity and variance of the noise.}

\textcolor{black}{
\begin{theorem}\label{thm:theorem2}
The required noise that must be added to a local model before sending it to the server in the first round is given by:
\begin{equation}
\vec{\Gamma}_{k}^{1} = \mathcal{N}\left(0, \frac {2D}{{\epsilon^2}} \ln{[1.25L]}\right)
\end{equation}
\end{theorem}
}
\textcolor{black}{
\begin{proofthm}\label{proof:proof1}
The noise \(\vec{\xi}_{k}^{1}\) is calculated based on Equation~\eqref{eq:sigma_dp}. For a specific value of \(\epsilon\), the sensitivity of the model is determined by Equation~\eqref{eq:sensitive}. A class hypervector is derived using Equation~\eqref{eq:formClassHypervector}. The difference between two class hypervectors, generated from two neighboring datasets, is itself a hypervector, implying that the sensitivity of a local model in the first round is the norm of this hypervector. Since hypervectors are binary, the sensitivity of the models is:
\begin{equation}
\Delta f = \sqrt{D}
\end{equation}
In the first round, each client uses \(L\) training samples to create the local model. An HD model consists of \(S\) class hypervectors. Therefore, \(L\) training samples contribute to generating \(S\) class hypervectors. In the worst-case scenario, all \(L\) samples belong to a single class. Consequently, for this class, \(L\) samples are used to form the model, and \(\delta\) is set as \(\frac{1}{L}\), since \(\delta\) is the inverse of the number of training samples. To account for the worst-case scenario where all training samples belong to one class, we set \(\delta\) as \(\frac{1}{L}\) for all classes. Thus, based on Equation~\eqref{eq:sigma_dp}, \(\vec{\xi}_{k}^{1}\) should be:
\begin{equation}
\vec{\xi}_{k}^{1} = \mathcal{N}\left(0,\frac {2D}{{\epsilon^2}} \ln{[1.25L]}\right)
\end{equation}
Since this is the first round:
\begin{equation}
\vec{\Gamma}_{k}^{1} = \vec{\xi}_{k}^{1}
\end{equation}
\end{proofthm}
}
\textcolor{black}{Once the local models are aggregated at the server, Theorem~\ref{thm:theorem3} establishes that no additional noise is required for the global model in the first round. This conclusion is demonstrated in Proof~\ref{proof:proof2}, which is further supported by Lemma~\ref{lem:lemma1} and Proof~\ref{proof:proof3}.}
\textcolor{black}{
\begin{theorem}\label{thm:theorem3}
Once the local models have been aggregated in the first round, the server does not need to add any additional noise to the global model. The noise introduced in the local models is sufficient to ensure the privacy of the global model.
\end{theorem}
}
\textcolor{black}{
\begin{proofthm}\label{proof:proof2}
A client sends a set of noisy class hypervectors to the server. Thus, the global model for class \(s\) in the first round is:
\begin{align}
\tilde{\vec{C}}_{g,1}^{s} &= \frac{\sum_{k=1}^{K} \tilde{\vec{C}}_{k,1}^{s}}{K} \\
&= \frac{\vec{C}_{1,1}^{s} + \vec{C}_{2,1}^{s} + \cdots + \vec{C}_{K,1}^{s}}{K} + \frac{\vec{\xi}_{1}^{1} + \vec{\xi}_{2}^{1} + \cdots + \vec{\xi}_{K}^{1}}{K} \notag
\end{align}
Here, $\vec{C}_{k,1}^{s}$ represents the class \(s\) hypervector for client \(k\), and \(\vec{\xi}_{k}^{1}\) is the noise added by client \(k\) to the class \(s\) hypervector at the first round. First, the noise in the global model can be defined as:
\begin{align}
\vec{\Psi}_{g}^{1} &= \frac{\vec{\xi}_{1}^{1} + \vec{\xi}_{2}^{1} + \cdots + \vec{\xi}_{K}^{1}}{K} \\
&= \frac{\mathcal{N}\left(0,\frac {2D}{{\epsilon^2}} \ln{[1.25L]}\right) + \mathcal{N}\left(0,\frac {2D}{{\epsilon^2}} \ln{[1.25L]}\right) + \cdots + \mathcal{N}\left(0,\frac {2D}{{\epsilon^2}} \ln{[1.25L]}\right)}{K} \notag
\end{align}
Since the noise terms are independent normal distributions:
\begin{align}
\vec{\Psi}_{g}^{1} &= \frac{\mathcal{N}\left(0, \frac{2DK}{{\epsilon^2}} \ln{[1.25L]}\right)}{K} \\
&= \mathcal{N}\left(0, \frac{2D}{{K\epsilon^2}} \ln{[1.25L]}\right) \notag
\end{align}
In this case, the signal part of the global model is $\vec{C}_{1,1}^{s} + \vec{C}_{2,1}^{s} +  \cdots + \vec{C}_{K,1}^{s}$. Each class hypervector is the summation of several hypervectors, so the sensitivity of the model can be expressed as:
\begin{align}
\Delta f &= \frac{\sqrt{D}}{K}
\end{align}
The server must secure the training samples from all clients for each class. In the worst-case scenario, where all clients use only the training samples from one class, the total number of training samples across clients is \(K \times L\), and \(\delta\) is defined as:
\begin{align}
\delta &= \frac{1}{KL}
\end{align}
Therefore, the required noise for the server to secure all the training samples from all clients, based on Equation~\eqref{eq:sigma_dp}, is:
\begin{align}
\vec{\xi}_{g}^{1} &= \mathcal{N}\left(0, \frac{2D}{K^2\epsilon^2} \ln{[1.25KL]}\right)
\end{align}
Now, define \(\gamma\) as:
\begin{align}
\gamma &\triangleq \frac{\text{Var}(\vec{\Psi}_{g}^{1})}{\text{Var}(\vec{\xi}_{g}^{1})} \\
&= \frac{\frac {2D}{{K\epsilon^2}} \ln{[1.25L]}}{\frac{2D}{K^2\epsilon^2} \ln{[1.25KL]}} \notag \\
&= \frac{K\ln{[1.25L]}}{\ln{[1.25KL]}} \notag
\end{align}
Since \(\gamma\) is always greater than one, the noise received from the clients is sufficient to secure the global model, and no additional noise needs to be added to the global model.
\end{proofthm}
}
\textcolor{black}{
\begin{lemma}\label{lem:lemma1}
For any integer values \(K\) and \(L\), where \(K \geq 2\) and \(L \geq 2\), the value of \(\gamma\) is greater than 1, where \(\gamma\) is defined as:
\begin{align}
\gamma &= \frac{K\ln{[1.25L]}}{\ln{[1.25KL]}}
\end{align}
\end{lemma}
}
\textcolor{black}{
\begin{proofthm}\label{proof:proof3}
To demonstrate that \(\gamma\) is an increasing function, the derivative of \(\gamma\) with respect to both \(K\) and \(L\) is calculated, and it is shown that both derivatives are greater than zero. The partial derivative of \(\gamma\) with respect to \(K\) is:
\begin{align}
\frac{\partial \gamma}{\partial K} &= \frac{\ln{[1.25L]}\ln{[1.25KL]}-K\left(\frac{1.25L}{1.25KL}\right)\ln{[1.25L]}}{\left(\ln{[1.25KL]}\right)^2} \\
&= \frac{\ln{[1.25L]}}{\left(\ln{[1.25KL]}\right)^2} \left( \ln{[1.25KL]} - 1\right)      \notag
\end{align}
Since \(K\) and \(L\) are greater than 2:
\begin{align}
\ln{[1.25L]} > 0 \\
\ln{[1.25KL]} > 1 
\end{align}
Therefore:
\begin{align}
\frac{\partial \gamma}{\partial K} > 0
\end{align}
The partial derivative of \(\gamma\) with respect to \(L\) is:
\begin{align}
\frac{\partial \gamma}{\partial L} &= \frac{K\left(\frac{1.25}{1.25L}\right)\ln{[1.25KL]}-K\left(\frac{1.25K}{1.25KL}\right)\ln{[1.25L]}}{\left(\ln{[1.25KL]}\right)^2} \\
&= \frac{\frac{K}{L}}{\left(\ln{[1.25KL]}\right)^2}\left( \ln{[1.25KL]} - \ln{[1.25L]} \right)        \notag
\end{align}
Since \(1.25KL > 1.25L\), it follows that:
\begin{align}
\ln{[1.25KL]} > \ln{[1.25L]}
\end{align}
Thus:
\begin{align}
\frac{\partial \gamma}{\partial L} > 0
\end{align}
Since both partial derivatives are positive, \(\gamma\) is an increasing function. The minimum value of \(\gamma\) occurs when \(K=2\) and \(L=2\):
\begin{align}
\gamma_{min} &= \frac{K\ln{[1.25L]}}{\ln{[1.25KL]}} \Bigg|_{K=2, L=2} \\
&= \frac{2\ln{[2.5]}}{\ln{[5]}} \notag  \\
&= 1.13 > 1  \notag
\end{align}
Therefore, \(\gamma\) is always greater than 1.
\end{proofthm}
}
\textcolor{black}{In continuous FL scenarios, new data is introduced in each round, necessitating noise adjustments to ensure cumulative privacy guarantees. For round \(r \geq 2\), the noise required for local models is defined in Theorem~\ref{thm:theorem4}, which considers both the cumulative noise from previous rounds and the additional noise needed for the current round, as demonstrated in Proof~\ref{proof:proof4}.}
\textcolor{black}{
\begin{theorem}\label{thm:theorem4}
The amount of noise that should be added to a local model before sending it to the server in round \(r\), where \(r \geq 2\), is given by:
\begin{equation}
\vec{\Gamma}_{k}^{r} = \mathcal{N}\left(0,\frac {2D}{{K\epsilon^2}} \left[K\ln{\left(1.25(r-1)KL + 1.25L\right)} - \ln{\left(1.25(r-2)KL + 1.25L\right)}\right]\right)
\end{equation}
\end{theorem}
}
\textcolor{black}{
\begin{proofthm}\label{proof:proof4}
The noise added by a client, \(\vec{\Gamma}_{k}^{r}\), is the difference between the required noise to protect all samples that contributed to the local model up to round \(r\), \(\vec{\xi}_{k}^{r}\), and the cumulative noise from previous rounds stored in the global model, \(\vec{\Psi}_{k}^{r-1}\). The noise \(\vec{\xi}_{k}^{r}\) is calculated based on Equation~\eqref{eq:sigma_dp}. For a specific value of \(\epsilon\), the sensitivity of the model is determined using Equation~\eqref{eq:sensitive}. In round \(r\) (where \(r \geq 2\)), a client only retrains the secured global model. When considering the system as a retraining of the HD model, the output difference for neighboring inputs is a hypervector. Since hypervectors are binary, the sensitivity of the system is:
\begin{equation}
\Delta f = \sqrt{D}
\end{equation}
In this scenario, each client downloads the global model from the server, which contains information from round \(r-1\), and uses \(L\) samples to retrain the global model. The total number of samples contributing to the global model up to round \(r-1\) is \(KL(r-1)\). Therefore, the total number of samples that need to be secured before sending the local model to the server in round \(r\) is \(KL(r-1) + L\). Thus, \(\delta\) is given by:
\begin{equation}
\delta = \frac{1}{KL(r-1) + L}
\end{equation}
Consequently, based on Equation~\eqref{eq:sigma_dp}, \(\vec{\xi}_{k}^{r}\) is:
\begin{equation}
\label{eq:req_client_r}
\vec{\xi}_{k}^{r} = \mathcal{N}\left(0, \frac{2D}{{\epsilon^2}} \ln{[1.25(r-1)KL + 1.25L]}\right)
\end{equation}
On the other hand, the global model retains signal information and noise from previous rounds. The global model does not add any additional noise; it merely averages the noisy local models. Therefore, \(\vec{\Psi}_{k}^{r-1}\) is the average noise from clients in round \(r-1\). In round \(r-1\), a client added sufficient noise to secure the training samples in round \(r-2\). Hence, we have:
\begin{align}
\vec{\Psi}_{k}^{r-1} &= \frac{\sum_{k=1}^{K} \vec{\xi}_{k}^{r-1}}{K} \\
&= \mathcal{N}(0, \frac{2D}{{K\epsilon^2}} \ln{[1.25(r-2)KL + 1.25L]}) \notag
\end{align}
Finally, we obtain:
\begin{align}
\vec{\Gamma}_{k}^{r} &= \vec{\xi}_{k}^{r} - \vec{\Psi}_{k}^{r-1} \\
&= \mathcal{N}(0, \frac{2D}{{K\epsilon^2}} \left[K\ln{[1.25(r-1)KL + 1.25L]} - \ln{[1.25(r-2)KL + 1.25L]}\right]) \notag
\end{align}
\end{proofthm}
}
\textcolor{black}{After aggregating the local models in round \(r \geq 2\), Theorem~\ref{thm:theorem5} establishes that no additional noise is required for the global model, as the privacy guarantees are maintained by the noise introduced in the local models. This result is rigorously demonstrated in Proof~\ref{proof:proof5}.}
\textcolor{black}{
\begin{theorem}\label{thm:theorem5}
After aggregating the local models in round \(r\), where \(r \geq 2\), the server does not need to add any additional noise to the global model since the noise introduced in the local models is sufficient to ensure the privacy of the global model.
\end{theorem}
}
\textcolor{black}{
\begin{proofthm}\label{proof:proof5}
At round \(r\), a client sends a set of noisy class hypervectors to the server. Thus, the global model for class \(s\) is given by:
\begin{align}
\tilde{\vec{C}}_{g,r}^{s} &= \frac{\sum_{k=1}^{K} \tilde{\vec{C}}_{k,r}^{s}}{K} \\
&= \frac{\vec{C}_{1,r}^{s} + \vec{C}_{2,r}^{s} + \cdots + \vec{C}_{K,r}^{s}}{K} + \frac{\vec{\xi}_{1}^{r} + \vec{\xi}_{2}^{r} + \cdots + \vec{\xi}_{K}^{r}}{K} \notag
\end{align}
where \(\vec{C}_{k,r}^{s}\) is the class \(s\) hypervector for client \(k\), and \(\vec{\xi}_{k}^{r}\) is the noise added to the class \(s\) hypervector up to round \(r\). First, the noise in the global model is defined as:
\begin{align}
\vec{\Psi}_{g}^{r} &= \frac{\vec{\xi}_{1}^{r} + \vec{\xi}_{2}^{r} + \cdots + \vec{\xi}_{K}^{r}}{K}\\
&= \frac{\mathcal{N}\left(0, \frac{2D}{{\epsilon^2}} \ln{\left[1.25(r-1)KL + 1.25L\right]}\right) + \cdots + \mathcal{N}\left(0, \frac{2D}{{\epsilon^2}} \ln{\left[1.25(r-1)KL + 1.25L\right]}\right)}{K} \notag
\end{align}
Since the normal distributions are independent:
\begin{align}
\label{eq:com_server_r}
\vec{\Psi}_{g}^{r} &= \frac{\mathcal{N}\left(0, \frac{2DK}{{\epsilon^2}} \ln{\left[1.25(r-1)KL + 1.25L\right]}\right)}{K}\\
&= \mathcal{N}\left(0, \frac{2D}{{K\epsilon^2}} \ln{\left[1.25(r-1)KL + 1.25L\right]}\right) \notag
\end{align}
In this case, the signal part of the global model is \(\vec{C}_{1,r}^{s} + \vec{C}_{2,r}^{s} + \cdots + \vec{C}_{K,r}^{s}\). Since each class hypervector is the sum of some hypervectors, the sensitivity of the model is defined as:
\begin{align}
\Delta f &= \frac{\sqrt{D}}{K}
\end{align}
The server must ensure the security of training samples from all clients for each class. Given that the total number of training samples from the clients is \(K \times L \times r\), \(\delta\) is defined as:
\begin{align}
\delta &= \frac{1}{KLr}
\end{align}
Thus, the required noise for the server to secure all training samples for all clients, based on Equation~\eqref{eq:sigma_dp}, is:
\begin{align}
\label{eq:req_server_r}
\vec{\xi}_{g}^{r} &= \mathcal{N}\left(0, \frac{2D}{K^2\epsilon^2} \ln{[1.25KLr]}\right)
\end{align}
Next, we define \(\gamma\) as:
\begin{align}
\gamma &\triangleq \frac{\text{Var}(\vec{\Psi}_{g}^{r})}{\text{Var}(\vec{\xi}_{g}^{r})} \\
&= \frac{\frac{2D}{{K\epsilon^2}} \ln{\left[1.25(r-1)KL + 1.25L\right]}}{\frac{2D}{K^2\epsilon^2} \ln{\left[1.25KLr\right]}} \notag \\
&= \frac{K\ln{\left[1.25(r-1)KL + 1.25L\right]}}{\ln{\left[1.25KLr\right]}} \notag
\end{align}
For any positive integer values of \(K\) and \(L\) greater than two, \(\gamma\) is always greater than one. Therefore, the amount of noise received from the clients is sufficient to ensure the privacy of the global model, and no additional noise needs to be added to the global model.
\end{proofthm}
}
\textcolor{black}{Lemma~\ref{lem:lemma2} ensures that the cumulative noise from clients is sufficient to protect the global model across rounds, eliminating the need for additional noise from the server. This result is further supported and demonstrated in Proof~\ref{proof:proof6}.}
\textcolor{black}{
\begin{lemma}\label{lem:lemma2}
For any positive integer values greater than 2 for K, L, and r, $\gamma$ is greater than 1, if we define:
\begin{align}
\gamma &= \frac{K\ln{[1.25(r-1)KL+1.25L]}}{\ln{[1.25KLr]}}
\end{align}
\end{lemma}
}
\textcolor{black}{
\begin{proofthm}\label{proof:proof6}
To demonstrate that \(\gamma\) is an increasing function, the derivative of \(\gamma\) with respect to \(K\), \(L\), and \(r\) is calculated, showing that all derivatives are greater than zero. First, define \(\beta\) as:
\begin{align}
\beta &\triangleq 1.25(r-1)KL+1.25L 
\end{align}
Then, the partial derivative of \(\gamma\) with respect to \(K\) is:
\begin{align}
\frac{\partial \gamma}{\partial K} &= \frac{\left( \ln{[\beta]}+K\left(\frac{1.25(r-1)L}{\beta}\right)\right)\ln{[1.25KLr]}-\frac{1.25Lr}{1.25KLr}\left( K\ln{[\beta]}\right)}{\left(\ln{[1.25KLr]}\right)^2} \\ 
&= \frac{\ln{[\beta]}\left( \ln{[1.25KLr]}-1 \right) + K\left(\frac{1.25(r-1)L}{\beta}\right)\ln{[1.25KLr]}}{\left(\ln{[1.25KLr]}\right)^2} \notag \\   
&= \frac{\ln[\beta]}{{\left(\ln{[1.25KLr]}\right)^2}}\left( \ln{[1.25KLr]}-1 \right) + \frac{K}{\ln{[1.25KLr]}}\left(\frac{1.25(r-1)L}{\beta}\right)      \notag
\end{align}
Since \(K\), \(L\), and \(r\) are greater than 2:
\begin{align}
\ln{[\beta]} > 0 \\
\ln{[1.25KLr]} > 1 
\end{align}
Thus:
\begin{align}
\frac{\partial \gamma}{\partial K} > 0
\end{align}
Now, for the partial derivative of \(\gamma\) with respect to \(r\):
\begin{align}
\frac{\partial \gamma}{\partial r} &= \frac{K\left(\frac{1.25KL}{\beta}\right)\ln{[1.25KLr]}-K\ln{[\beta]}\frac{1.25KL}{1.25KLr}}{\left(\ln{[1.25KLr]}\right)^2} \\ 
 &=  \frac{K}{\left(\ln{[1.25KLr]}\right)^2}  \left( \frac{1.25KL}{\beta}\ln{[1.25KLr]} - \frac{\ln{[\beta]}}{r}  
    \right)              \notag
\end{align}
We know \(\beta\) is:
\begin{align}
\beta &= 1.25(r-1)KL+1.25L \\
    &= 1.25KLr - 1.25KL + 1.25L \notag
\end{align}
Since \(1.25KL\) is greater than \(1.25L\):
\begin{align}
\beta &< 1.25KLr 
\end{align}
Thus:
\begin{align}
\ln{[1.25KLr]} &> \ln{[\beta]} 
\label{eq:lem61}
\end{align}
Additionally:
\begin{align}
\frac{1.25KL}{\beta} &> \frac{1.25KL}{1.25KLr} = \frac{1}{r} 
\end{align}
Therefore:
\begin{align}
\frac{1.25KL}{\beta}\ln{[1.25KLr]} &> \frac{\ln{[\beta]}}{r} 
\end{align}
And:
\begin{align}
\frac{\partial \gamma}{\partial r} > 0
\end{align}
Next, the partial derivative of \(\gamma\) with respect to \(L\) is:
\begin{align}
\frac{\partial \gamma}{\partial L} &= \frac{K\left(  \frac{1.25(r-1)K+1.25}{\beta}\right)\ln{[1.25KLr]} -\left( \frac{1.25Kr}{1.25KLr} \right)K\ln{[\beta]}}{\left(\ln{[1.25KLr]}\right)^2} \\
&=    \frac{K}{\left(\ln{[1.25KLr]}\right)^2}  \left(   \frac{1.25(r-1)K+1.25}{\beta}\ln{[1.25KLr]} - \frac{\ln{[\beta]}}{L} \right)         \notag  \\
&=  \frac{K}{\left(\ln{[1.25KLr]}\right)^2} \left( \frac{\ln{[1.25KLr]}-\ln[\beta]}{L}\right)  \notag  
\end{align}
From Equation~\eqref{eq:lem61}:
\begin{align}
\frac{\partial \gamma}{\partial L} > 0
\end{align}
Since all partial derivatives are positive, \(\gamma\) is an increasing function. The minimum value of \(\gamma\) occurs when \(K=2\), \(L=2\), and \(r=2\):
\begin{align}
\gamma &= \frac{K\ln{[1.25(r-1)KL+1.25L]}}{\ln{1.25KLr}} \Bigg|_{K=2, L=2, r=2} \\
&= \frac{2\ln{[7.5]}}{\ln{[10]}} \notag  \\
&= 1.75 > 1  \notag
\end{align}
Therefore, \(\gamma\) is always greater than 1.
\end{proofthm}
}

\begin{figure}
    \centering
    \includegraphics[width=\textwidth]{ 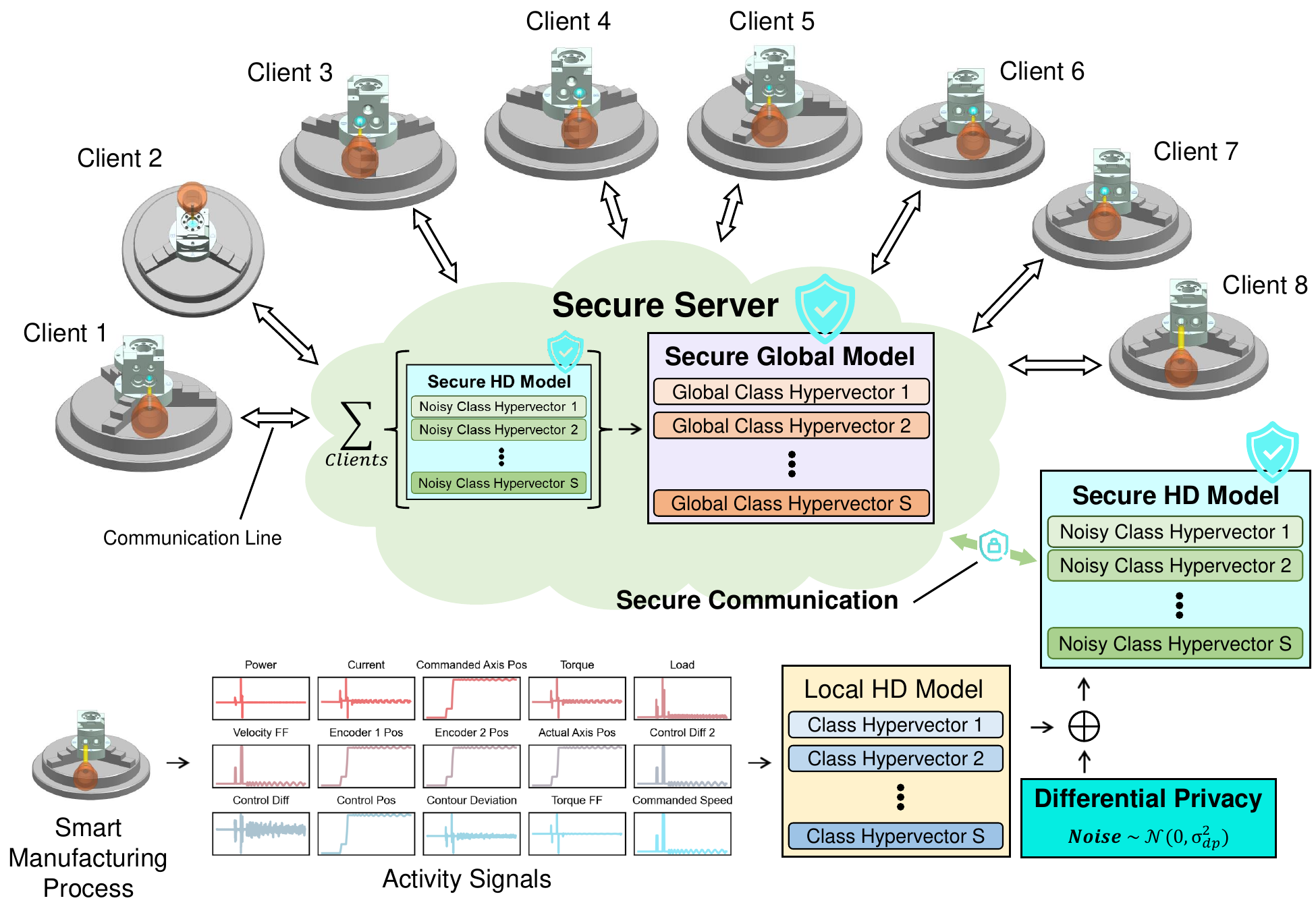}
        \caption{Experimental setup for federated learning in smart manufacturing. The federated learning structure applied to a smart manufacturing environment with 8 clients, each operating independently to perform quality control tasks. Each client collects local data, processes it, and updates local models, which are then aggregated into a global model to enhance accuracy and maintain data privacy across the network.}
    \label{fig:dataset}
\end{figure}
\section{Experimental Design}
\label{sec:Experimental Design}

In industrial IoT systems, vast amounts of data are generated and require
advanced analytical tools such as AI and ML to drive efficiency and improve
production quality within smart manufacturing
environments~\cite{praveena2022comprehensive}. The geometric characteristics of
manufactured parts, particularly their precision, play a crucial role in
determining mechanical properties like stiffness, strength, and failure
resistance~\cite{ramirez2023review}. Research has shown that even minor
deviations from intended dimensions can result in defects, potentially leading
to part failure~\cite{naranjo2019tensile}. Thus, accurate measurement,
prediction, and control over these geometric dimensions are essential to
maintain high product quality and prevent production
defects~\cite{fu2022machine}.

\subsection{\textcolor{black}{Datasets and Data Preprocessing}}

In our case study, we implemented FL for quality control in smart manufacturing, focusing on 8 clients, as shown in  Figure~\ref{fig:dataset}. Each client used a LASERTEC 65 DED hybrid CNC DMG machine to perform drilling tasks, producing 18 parts from 1040 steel blocks with dimensions of 76.2~mm $\times$ 76.2~mm $\times$ 76.2~mm. The diameter of each drilled hole was measured, and Z-scores were calculated to categorize parts into three groups: \textit{Nominal Drilling} ($-1 \leq Z \leq 1$), \textit{Under Drilling} ($Z < -1$), and \textit{Over Drilling} ($Z > 1$).

Real-time data was collected using a Siemens Simatic IPC227E device connected to the DMG machine. This device recorded 15 critical process signals—including Load, Current, Torque, Commanded Speed, Control Differential (Diff), Power, Contour Deviation, Encoder Positions, Velocity Feed Forward (FF), and Torque FF—at a frequency of 500 Hz during the manufacturing process. These signals, captured across five axes and the spindle, totaled 90 process signals per part.

\begin{table}[h!]
\centering
\color{black}

\caption{\textcolor{black}{Summary of transformer and convolutional neural network architectures}}
\label{table:bench_architecture}
\begin{tabular}{c c}
\hline
\textbf{Model}        & \textbf{Details}                                                                                     \\ \hline

\multirow{4}{*}{Transformer} 
                      & Input Projection: Linear layer (10 → 128)                                                             \\
                      & Positional Encoding: Learnable embeddings                                                             \\
                      & 2 Transformer Encoder layers: 4 heads, FFN size 256, ReLU, Dropout=0.1                                \\
                      & 2 Fully Connected layers: 128 → 8                                                                     \\ \hline
\multirow{3}{*}{Convolutional Neural Network}  & 2 Convolutional layers: 32 and 64 filters (3x3 kernels, ReLU activation)                              \\
                      & MaxPooling layer (2x2, stride 2)                                                                      \\
                      & 3 Fully Connected layers: 14400 → 1024 → 1024 → 8                                                    \\ \hline
\end{tabular}
\end{table}

To generate samples, each part’s 90 recorded signals were segmented into subsignals using a sliding window approach with a window size of 10. This resulted in multiple samples per part, where each sample contained 90 signals of length 10. The label assigned to each sample was inherited from the Z-score category of the part from which the data was recorded.

\subsection{\textcolor{black}{Experimental Setup}}
For balanced evaluation, we ensured equal representation from each label category (\textit{Nominal Drilling}, \textit{Under Drilling}, and \textit{Over Drilling}) by selecting an equal number of samples per client. The dataset was split such that 80\% of samples were used for training and 20\% for testing the global model. During FL training, each client's training set was divided into 10 subsets. In each communication round, a single subset was used to update the local model, ensuring 10 communication rounds in total. The test dataset was constructed by combining all clients' test samples. The global model's performance was evaluated on the aggregated test data, ensuring robust and accurate results.

\subsection{\textcolor{black}{Hyperparameters and Evaluation Metrics}}
We conducted a sensitivity analysis to evaluate the proposed method under different settings by tuning key hyperparameters. Noise variance was adjusted based on the privacy budget ($\epsilon$) and tracked across rounds to minimize cumulative effects. The hypervector size and privacy budget ($\epsilon$) were varied to examine their influence on model accuracy, training time, and energy consumption. The number of rounds was fixed at 10 to balance communication overhead and model convergence. To assess model performance, we used accuracy, which measures the proportion of correctly classified samples in the test set; training time, which represents the total duration required for model convergence; and energy consumption, which was measured at the device level during training to evaluate efficiency.

\subsection{\textcolor{black}{Comparison with Benchmark Frameworks}}

We compared FedHDPrivacy against FedAvg, FedNova, FedProx, and FedOpt. These frameworks were trained using two ML architectures: a Transformer and a Convolutional Neural Network (CNN), detailed in Table~\ref{table:bench_architecture}. Each benchmark was tested under identical conditions using the same dataset and experimental setup. The performance of these frameworks was analyzed across the selected metrics to highlight the advantages of the proposed method in terms of privacy preservation, accuracy, and efficiency.

\section{Experimental Results}
\label{sec:Experimental Results}

 \begin{figure}
    \centering
    \includegraphics[width=\textwidth]{ 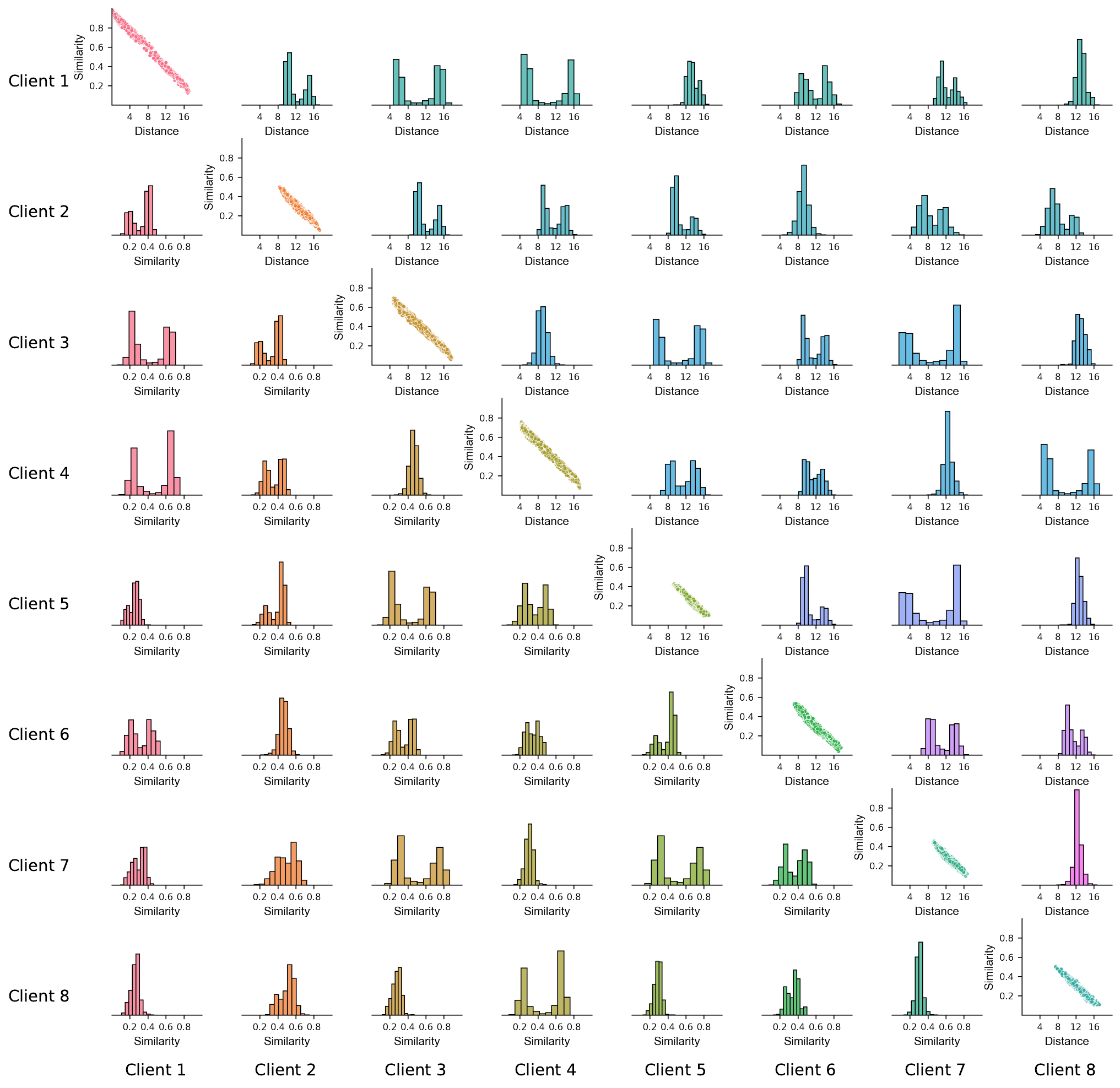}
    \caption{Distance and similarity analysis in hyperdimensional encoding. A visualization of distance and similarity relationships in client data. The main diagonal shows the similarity of encoded samples as a function of distance within each client. The lower triangle displays the similarity distribution between clients, while the upper triangle shows the distance distribution between raw data samples across clients. Together, these elements reveal patterns in data structure, client diversity, and cross-client similarity in the system.}
    \label{fig:dis_sig_basis}
\end{figure}

\subsection{{\textcolor{black}{Sensitivity Analysis}}}

Figure~\ref{fig:dis_sig_basis} illustrates the relationship between distance and similarity among clients' data points. The main diagonal of the figure presents the similarity between encoded samples as a function of the distance between them for each individual client. This provides valuable insight into how well the encoding represents the internal structure of the data for each client. A high similarity for samples with shorter distances suggests that the encoding method is successfully capturing the inherent relationships within the client’s data. Notably, the data points for each client align on a line, indicating that the encoding strategy maps the data into the hyperspace effectively, without any outliers. For certain clients, such as Client 1, the similarity values range from 0 to 1, indicating a wide diversity between samples. In contrast, for Client 5, the similarity ranges from 0 to approximately 0.4, while the distances vary only between 8 and 16. This pattern suggests that Client 5's samples are quite diverse, with little overlap in similarity, which highlights the distinctiveness of the data within this client's set.

The lower triangle of Figure~\ref{fig:dis_sig_basis} shows the distribution of similarity between the hypervectors (encoded samples) for each pair of clients. This comparison highlights the level of similarity in the data representations across different clients. A higher similarity between clients may indicate commonality in the data characteristics, which could be relevant for tasks such as global model aggregation in FL. For instance, Clients 5 and 7 exhibit greater similarity, as their distribution peaks around a similarity value of 0.8, despite not having high similarity between the hypervectors of individual samples. In contrast, Clients 4 and 3 appear to have less similarity between their data, even though their individual samples, as shown in the main diagonal, exhibit similar internal patterns. This difference in cross-client similarity highlights the variation in data structure between different clients.

In contrast, the upper triangle of Figure~\ref{fig:dis_sig_basis} presents the distribution of distances between the actual samples from each pair of clients. This provides insight into the divergence or diversity in the raw data across clients. Large distances between samples suggest significant differences in their data distributions, which may affect the model’s ability to generalize effectively across all clients. This combination of encoded sample similarities and raw data distances offers a holistic view of both the encoded representations and the underlying diversity in the system. For example, Clients 1 and 8 exhibit high distance values, with a distribution ranging from 12 to 16, indicating that the raw data for these clients is notably different. On the other hand, Clients 3 and 4 have more similar raw data points, as evidenced by a distance distribution peaking around 8.

 \begin{figure}
    \centering
    \includegraphics[width=\textwidth]{ 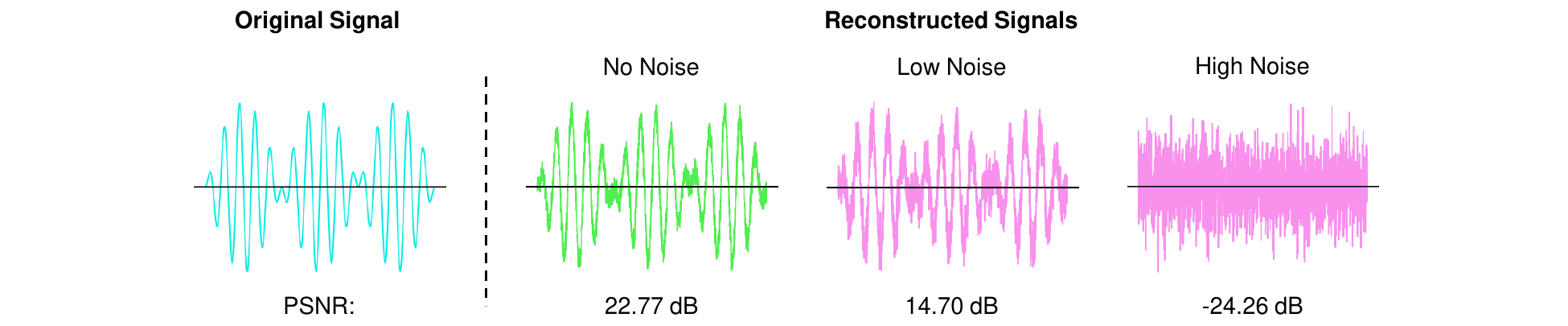}
        \caption{Signal reconstruction from hypervectors under varying noise levels. Comparison of original and reconstructed signals under conditions of no noise, low noise, and high noise levels.}
    \label{fig:noisy_signals}
\end{figure}

\begin{figure}
    \centering
    \includegraphics[width=\textwidth]{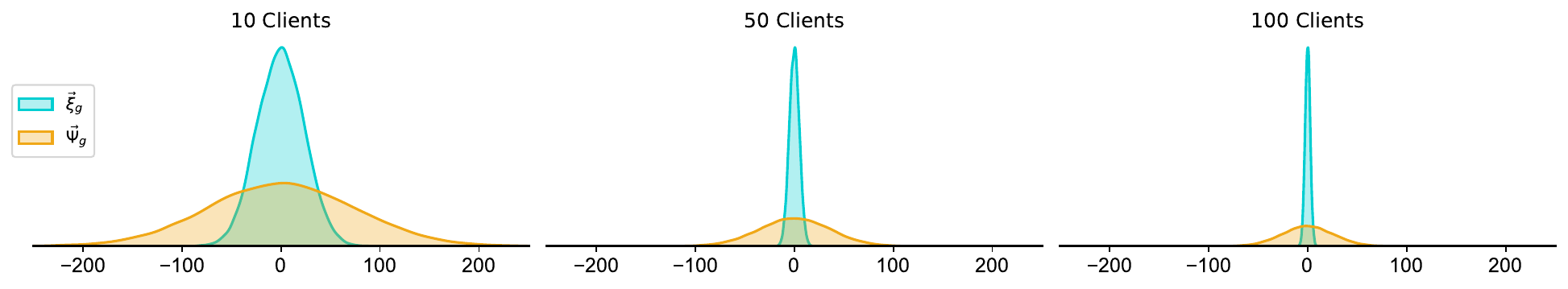}
        \caption{Cumulative and required noise levels for the server's model. FedHDPrivacy framework noise levels for server models with 10, 50, and 100 clients, illustrating cumulative versus required noise. The framework controls noise per communication round, ensuring privacy without needing server-side noise. As the client count rises, cumulative noise variance decreases, balancing privacy with accuracy across rounds.}
    \label{fig:noise_ser_k}
\end{figure}

 \begin{figure}
    \centering
    \begin{subfigure}[b]{0.6\textwidth}
        \centering
        \includegraphics[width=\textwidth]{ 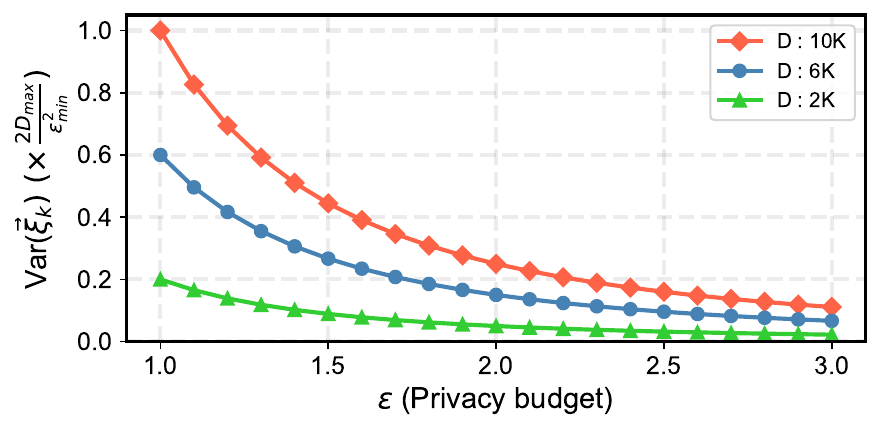}
        \caption{}
        \label{fig:noise_3d_eps}
    \end{subfigure}
    \hfill
    \begin{subfigure}[b]{0.6\textwidth}
        \centering
        \includegraphics[width=\textwidth]{ 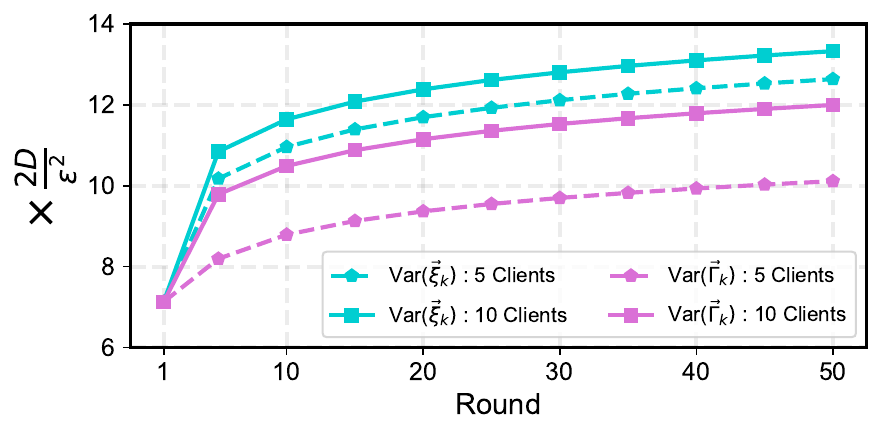}
        \caption{}
        \label{fig:noise_rnd_clt}
    \end{subfigure}
    \caption{Required and added noise for clients' models. (a) Relationship between required noise, hypervector size, and privacy budget. (b) Comparison of required noise and the added noise, which represents the difference between required noise and cumulative noise, across communication rounds.}
    \label{fig:noise_eps_priv}
\end{figure}

\begin{figure}
    \centering
    \includegraphics[width=0.6\textwidth]{ 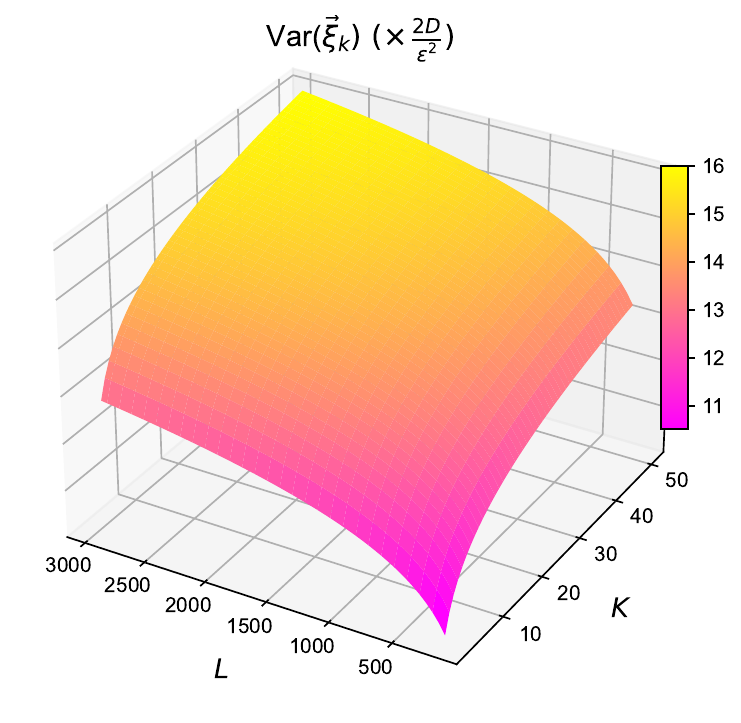}
        \caption{Required noise as a function of the number of clients and the quantity of training samples. The required noise for local models increases as both the number of clients and the quantity of training samples grow.}
    \label{fig:noise_3d}
\end{figure}

A HD model, as an XAI model, is vulnerable to revealing information about its training samples. For instance, after encoding, a signal can be decoded to reconstruct the original signal. Figure~\ref{fig:noisy_signals} demonstrates that the signal can be reconstructed with a Peak Signal-to-Noise Ratio (PSNR) of $22.77$ dB, showing that without adequate protection, the original signal can be easily recovered. To prevent this reconstruction and secure the information, noise needs to be added. However, the amount of noise directly impacts the privacy level. For example, Figure~\ref{fig:noisy_signals} also shows that a noisy encoded signal with a low level of noise can still be reconstructed with a PSNR of $14.70$ dB, where the overall behavior of the signal remains visible. On the other hand, when a higher level of noise is introduced, the reconstructed signal becomes unrecognizable, as shown by a PSNR of $-24.26$ dB, indicating that the reconstruction is no longer related to the original signal. However, adding too much noise can negatively impact the model’s performance. Therefore, it is crucial to strike a balance between maintaining privacy and ensuring accuracy.

\begin{figure}
    \centering
    \includegraphics[width=\textwidth]{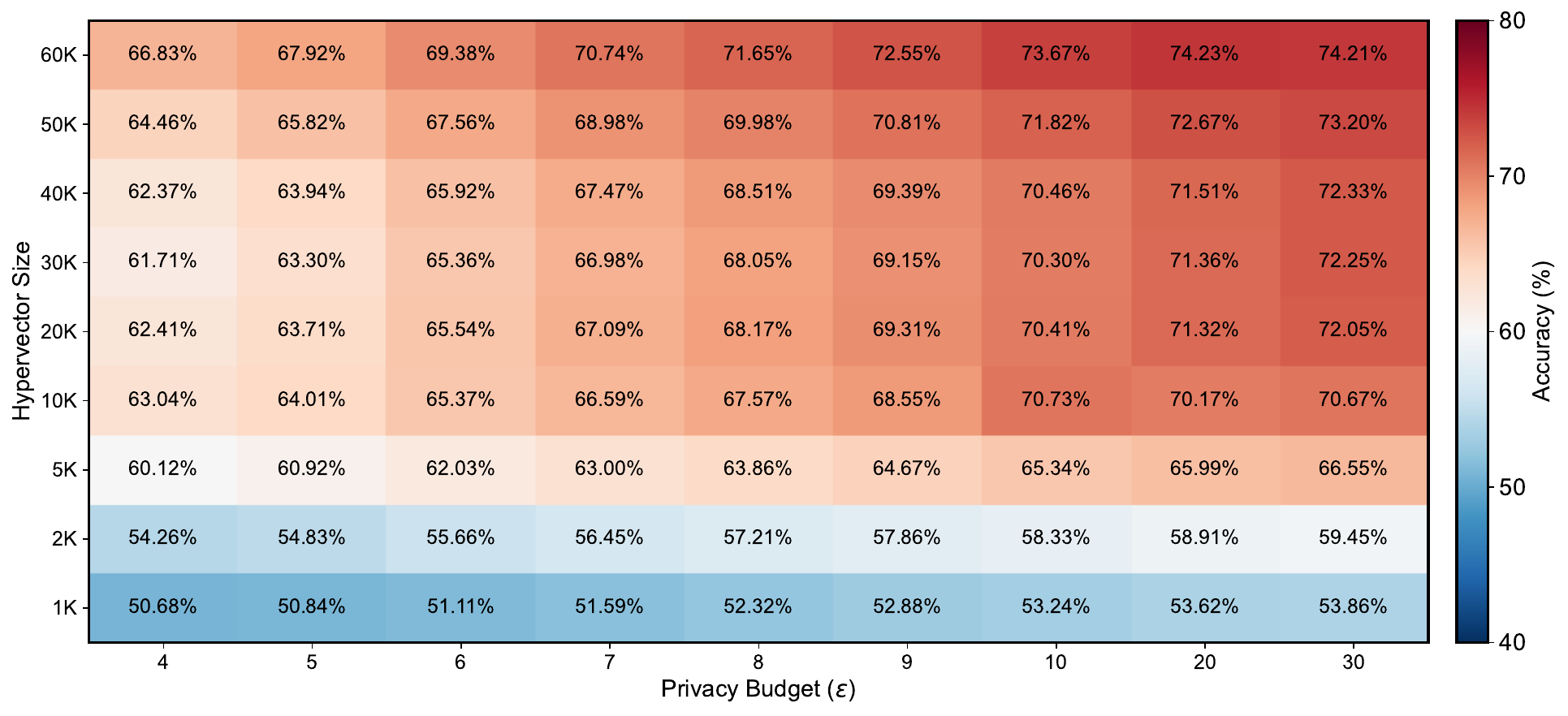}
        \caption{\textcolor{black}{Accuracy of FedHDPrivacy as a function of hypervector size and privacy budget. Larger hypervector sizes improve accuracy by increasing the model's capacity, while lower privacy budgets ($\epsilon$), indicating higher privacy levels, reduce accuracy.}}
    \label{fig:acc_heatmap}
\end{figure}

\begin{figure}
    \centering
    \includegraphics[width=0.6\textwidth]{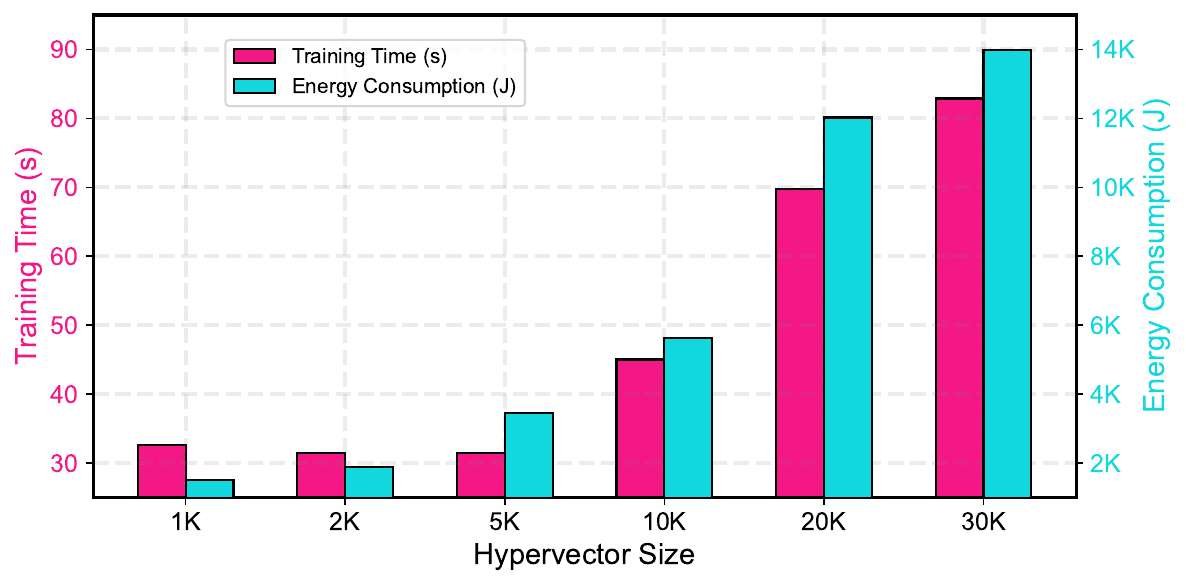}
    \caption{\textcolor{black}{Energy efficiency and training time as a function of hypervector size. Larger hypervector sizes require more energy and time for the training process.}}
    \label{fig:energy_traintime_barchart}
\end{figure}

To effectively balance privacy and accuracy, the FedHDPrivacy framework controls the amount of noise added in each communication round. We first demonstrated that the server's global model remains secure without the need for additional noise at the server side. Figure~\ref{fig:noise_ser_k} illustrates the cumulative noise and the required noise for the global model after 50 communication rounds. The results show that the variance of the cumulative noise consistently exceeds that of the required noise for any given number of clients. Additionally, as the number of clients increases from 10 to 100, the variances of both the required noise and the cumulative noise decrease, which can be attributed to the inverse relationship between noise and the number of clients, as established in Equation~\eqref{eq:com_server_r} and Equation~\eqref{eq:req_server_r}. Since the global model aggregates local models from all clients, an increase in the number of clients allows more noise to be canceled out, resulting in a lower cumulative noise. Moreover, the sensitivity of the global model also decreases as the number of clients increases, further reducing the required noise. Therefore, while cumulative noise is always greater than required noise, adding additional noise to the server is unnecessary, as the inherent structure of FedHDPrivacy ensures robust privacy protection.

For the clients' models, FedHDPrivacy accurately calculates the cumulative noise from previous communication rounds, the required noise for the current round, and finally, only adds the necessary amount to ensure security. The required noise has an inverse relationship with the privacy budget; as the privacy budget decreases, the noise level increases, and vice versa, as shown in Equation~\eqref{eq:req_client_r}. Additionally, the size of the hyperspace is directly proportional to the required noise, meaning larger hyperspaces demand higher levels of noise to maintain privacy, as outlined in Equation~\eqref{eq:req_client_r}. Figure~\ref{fig:noise_3d_eps} demonstrates that the normalized required noise for local models decreases with an increasing privacy budget, corresponding to lower noise levels. However, as the size of the hypervectors increases, the required noise rises accordingly, since larger hyperspaces necessitate more noise to protect the local models.

The required noise for local models also increases as the number of clients and training samples per round grows, as depicted in Figure~\ref{fig:noise_3d}. Given the dynamic nature of IoT systems, local models must be updated continuously over time. However, as shown in Figure~\ref{fig:noise_rnd_clt}, the required noise increases with each communication round. This occurs because, with each round, more cumulative training data from previous rounds is used to update the local models, necessitating more noise to secure the models. FedHDPrivacy addresses this by calculating the cumulative noise from previous rounds and only adding the difference between the required and cumulative noise for each round. Figure~\ref{fig:noise_rnd_clt} illustrates the normalized noise levels for local models over 50 communication rounds. Notably, the added noise is consistently lower than the required noise. For instance, after 50 rounds, the added noise is 80.03\% and 90\% of the required noise in the scenarios with 5 and 10 clients, respectively. This results in a reduction of 19.97\% and 10\% in noise added by the clients for the 5- and 10-client scenarios, respectively.

\textcolor{black}{For the given task with a fixed number of clients and training samples, the main hyperparameters of the proposed method are hypervector size and privacy budget. Figure~\ref{fig:acc_heatmap} illustrates the accuracy of FedHDPrivacy as a function of hypervector size and privacy budget. Larger hypervector sizes improve accuracy by increasing the model’s capacity, while lower privacy budgets (\(\epsilon\)), which indicate higher privacy levels, reduce accuracy. When the hypervector size is 1K, the accuracy is below 54\%, even with low noise levels, demonstrating that the model requires a larger hyperspace dimension for effective training. By increasing the hypervector size, accuracy improves significantly. For example, when the hypervector size is 10K, the accuracy exceeds 63\%, even at high privacy levels. Furthermore, decreasing the privacy level (i.e., increasing \(\epsilon\)) reduces the noise, thereby increasing the model's accuracy. For instance, when \(\epsilon = 10\), the accuracy surpasses 70\%. Finally, at a hypervector size of 60K, the accuracy improves further. In this case, by applying the minimum amount of noise, the accuracy reaches 74.21\%. Thus, increasing the hypervector size enhances accuracy.}

\textcolor{black}{However, increasing the hypervector size also increases training time and energy consumption, as shown in Figure~\ref{fig:energy_traintime_barchart}. For example, when the hypervector size is 1K, the training time and energy consumption are 32.56 seconds and 1.5 KJ, respectively. By increasing the hypervector size to 30K, the training time and energy consumption rise to 82.88 seconds and 13.98 KJ, respectively. This indicates that training time increases by 154.55\% and energy consumption increases by 832\% as the hypervector size grows. To balance accuracy, privacy, and efficiency, we selected a hypervector size of 10K and set \(\epsilon = 10\) for benchmarking against other FL frameworks.}

 \begin{figure}
    \centering
    \begin{subfigure}[b]{0.6\textwidth}
        \centering
        \includegraphics[width=\textwidth]{ 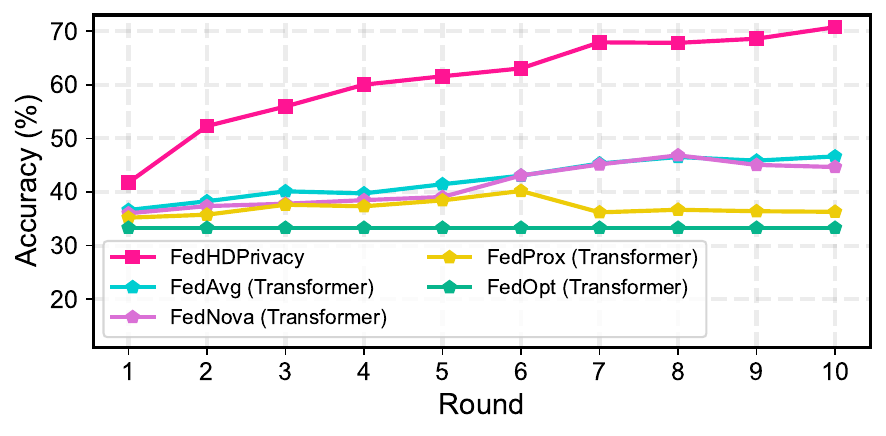}
        \caption{}
        \label{fig:acc_bench_tran}
    \end{subfigure}
    \hfill
    \begin{subfigure}[b]{0.6\textwidth}
        \centering
        \includegraphics[width=\textwidth]{ 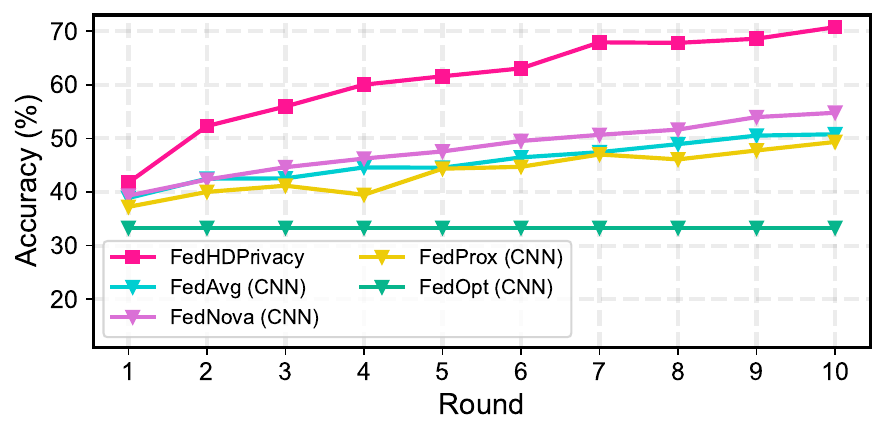}
        \caption{}
        \label{fig:acc_bench_cnn}
    \end{subfigure}
    \caption{\textcolor{black}{Accuracy comparison of FedHDPrivacy and benchmark FL frameworks using (a) transformer and (b) convolutional neural network over communication rounds. FedHDPrivacy outperforms FedAvg, FedNova, FedProx, and FedOpt by effectively managing noise to maintain accuracy across communication rounds, demonstrating superior performance in continuous learning scenarios.}}
    \label{fig:acc_bench}
\end{figure}

\begin{table}[h!]
\centering
\color{black}

\caption{\textcolor{black}{Comparison of false positive and false negative rates across different models.}}
\begin{tabular}{c c c c}
\hline
Federated Learning Framework & Machine Learning Model & False Positive Rate $\downarrow$ & False Negative Rate $\downarrow$ \\
\hline
FedHDPrivacy & Hyperdimensional Computing & \textbf{14.63\%} & \textbf{29.27\%} \\
FedAvg~\cite{mcmahan2017communication} & Transformer & 26.68\% & 53.36\% \\
FedNova~\cite{wang2020federated} & Transformer & 27.67\% & 55.34\% \\
FedProx~\cite{li2020federated} & Transformer & 31.85\% & 63.69\% \\
FedOpt~\cite{reddi2020adaptive} & Transformer & 33.33\% & 66.67\% \\
FedAvg~\cite{mcmahan2017communication} & Convolutional Neural Network & 24.62\% & 49.23\% \\
FedNova~\cite{wang2020federated} & Convolutional Neural Network & 22.61\% & 45.22\% \\
FedProx~\cite{li2020federated} & Convolutional Neural Network & 25.33\% & 50.66\% \\
FedOpt~\cite{reddi2020adaptive} & Convolutional Neural Network & 33.33\% & 66.67\% \\

\hline
\end{tabular}
\label{tab:fnr_fpr_table}
\end{table}

\subsection{\textcolor{black}{Benchmark Comparison}}

\textcolor{black}{Figure~\ref{fig:acc_bench} presents the accuracy comparison between the proposed method, FedHDPrivacy, and other FL frameworks. We benchmarked our framework against FedAvg, FedNova, FedProx, and FedOpt. Figure~\ref{fig:acc_bench_tran} illustrates the comparison with FL frameworks integrated with a transformer over communication rounds. In a continuous learning scenario spanning 10 communication rounds, the accuracy achieved by FedAvg, FedNova, FedProx, and FedOpt was $46.64\%$, $44.65\%$, $36.31\%$, and $33.32\%$, respectively—all falling below $50\%$. Notably, both FedProx and FedOpt performed poorly, with accuracy below $37\%$, highlighting their limited ability to distinguish between the three classes in the experimental task. In contrast, FedHDPrivacy, with a privacy budget $\epsilon = 10$, achieved an accuracy of $70.73\%$ after 10 rounds, significantly outperforming all benchmarked frameworks.}

\begin{figure}
    \centering
    \includegraphics[width=\textwidth]{ 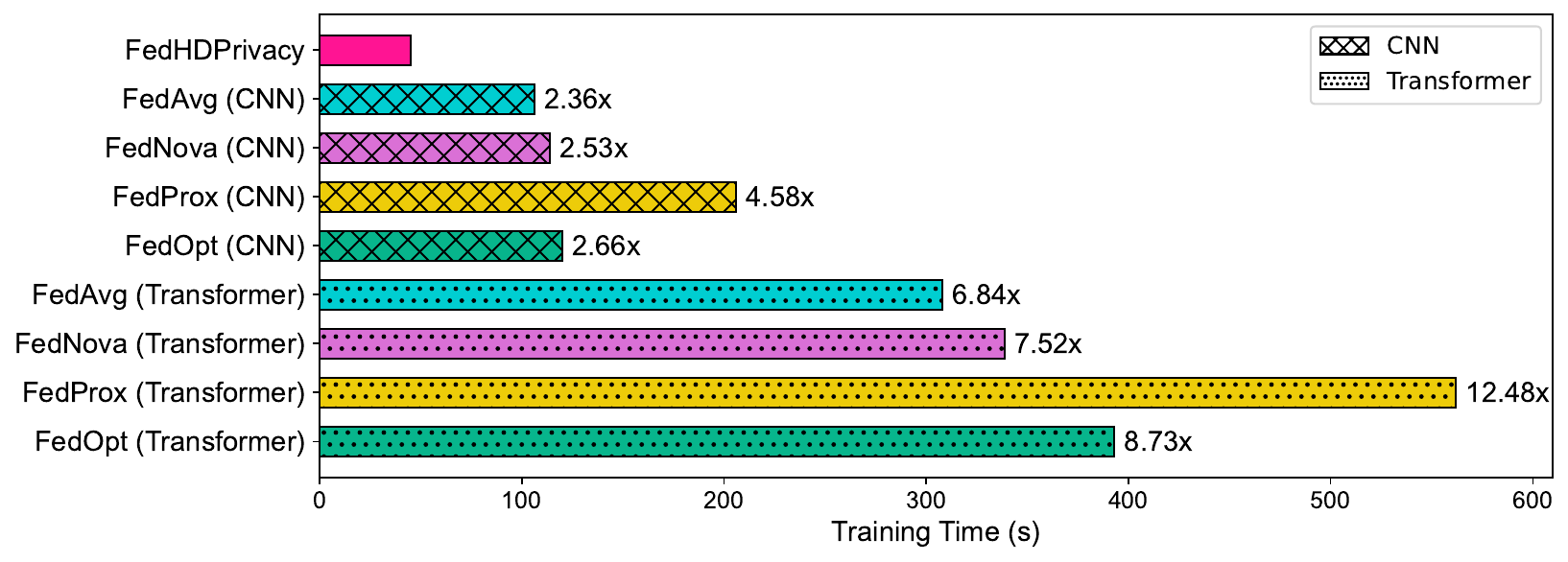}
        \caption{\textcolor{black}{Training time comparison of FedHDPrivacy and benchmark FL frameworks using a transformer and a convolutional neural network. FedHDPrivacy achieves faster training compared to FedAvg, FedNova, FedProx, and FedOpt.}}
    \label{fig:bench_traintime}
\end{figure}

\textcolor{black}{While the initial accuracy of all frameworks ranged between $33\%$ and $42\%$, the accuracy of FedHDPrivacy steadily increased with each communication round. For example, FedHDPrivacy's accuracy improved to $52.29\%$, $60.00\%$, $63.04\%$, $67.80\%$, and $70.73\%$ after 2, 4, 6, 8, and 10 rounds, respectively. This steady improvement highlights the superior performance of the proposed framework in continuous learning scenarios. Moreover, FedHDPrivacy demonstrated an overall accuracy improvement of $28.98\%$ after 10 rounds, significantly exceeding the best improvement seen in the other frameworks, which was less than $11\%$. For instance, FedNova's accuracy only increased from $36.02\%$ to $44.65\%$. This trend underscores FedHDPrivacy's ability to achieve consistent and notable accuracy gains over communication rounds.}

\textcolor{black}{Additionally, Figure~\ref{fig:acc_bench_cnn} depicts the comparison with FL frameworks integrated with a CNN over communication rounds. Once again, FedHDPrivacy consistently outperformed all other frameworks. While FedNova exhibited the closest performance to the proposed method, its accuracy remained significantly lower. For example, FedNova's accuracy was $9.97\%$, $13.77\%$, $13.53\%$, $16.16\%$, and $15.95\%$ lower than FedHDPrivacy after 2, 4, 6, 8, and 10 rounds, respectively. These results demonstrate the superior accuracy of FedHDPrivacy compared to benchmarked frameworks in both transformer-based and CNN-based FL settings.}

\textcolor{black}{Besides accuracy, we also evaluate the False Positive Rate (FPR) and False Negative Rate (FNR) for FedHDPrivacy and other benchmarks after the training process, as presented in Table~\ref{tab:fnr_fpr_table}. These metrics provide insights into the mistakes made by the models in incorrect predictions. The FPR for FedHDPrivacy is 14.63\%, while the best transformer-based benchmark, FedAvg, shows a significantly higher FPR of 26.68\%, highlighting the limitations of benchmark frameworks in accurate predictions. Similarly, the best CNN-based FL framework, FedNova, achieves an FPR of 22.61\%, which is nearly double that of FedHDPrivacy.}

\textcolor{black}{For the FNR, FedHDPrivacy achieves a value of 29.27\%, significantly lower than the best CNN-based benchmark, FedNova, which has an FNR of 45.22\%. Notably, the worst-performing framework, FedOpt, exhibits an FNR of 66.67\%, more than double the FNR of FedHDPrivacy. These results emphasize the superior ability of FedHDPrivacy to minimize incorrect predictions compared to the benchmark frameworks.}

\textcolor{black}{Figure~\ref{fig:bench_traintime} compares the training time of FedHDPrivacy with other FL frameworks. The training time for FedHDPrivacy is 45.02 seconds. In contrast, the training time for other FL frameworks is at least 2.36 times that of the proposed method, as seen in the case of the CNN-based FedAvg. The training time for CNN-based models is generally lower than transformer-based models due to the computational overhead of transformers. Notably, the training time for transformer-based models is up to 12.48 times that of FedHDPrivacy, as observed for FedProx, highlighting the efficiency and speed of the proposed method's training procedure.}

\textcolor{black}{Energy efficiency is a critical aspect of FL frameworks, particularly in IoT systems where devices often operate under strict energy constraints. Since the training process primarily occurs on the clients, we evaluate the energy consumption of clients across different FL frameworks while considering both transformer- and CNN-based architectures. }

\begin{figure}
    \centering
    \begin{subfigure}[b]{0.49\textwidth}
        \centering
        \includegraphics[width=\textwidth]{ 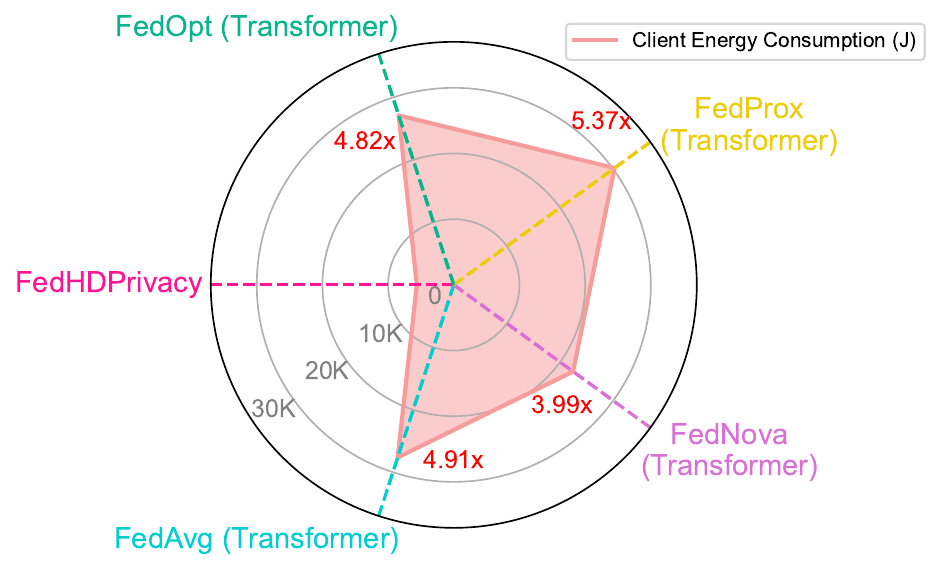}
        \caption{}
        \label{fig:bench_clientenergy1}
    \end{subfigure}
    \hfill
    \begin{subfigure}[b]{0.49\textwidth}
        \centering
        \includegraphics[width=\textwidth]{ 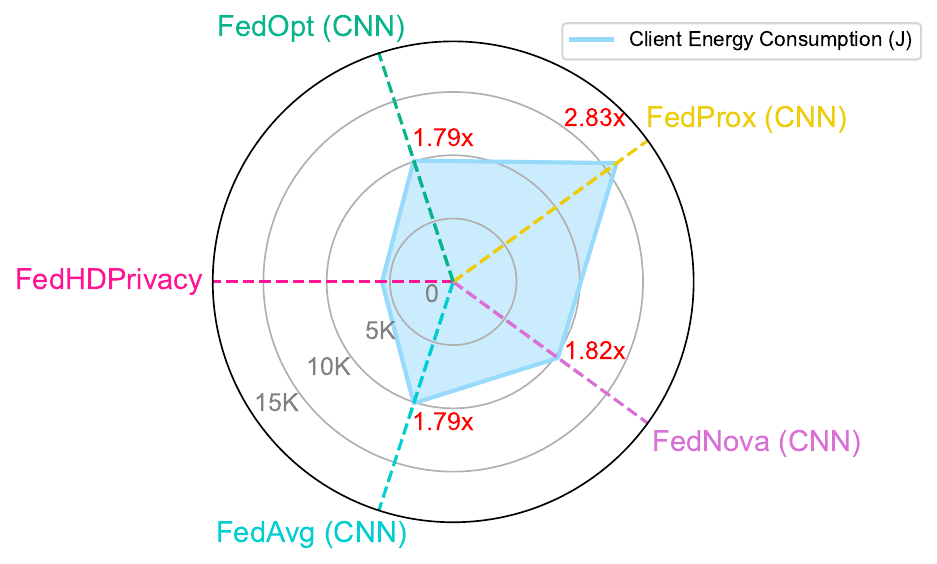}
        \caption{}
        \label{fig:bench_clientenergy2}
    \end{subfigure}
    \caption{\textcolor{black}{Client energy consumption comparison of FedHDPrivacy and benchmark FL frameworks using (a) transformer and (b) convolutional neural network. FedHDPrivacy demonstrates superior energy efficiency, consuming less energy compared to FedAvg, FedNova, FedProx, and FedOpt.}}
    \label{fig:bench_clientenergy}
\end{figure}

\textcolor{black}{Figure~\ref{fig:bench_clientenergy1} highlights the energy consumption of FedHDPrivacy compared to transformer-based FL frameworks. A single client in the FedHDPrivacy framework consumes 5.62 KJ, showcasing its energy-efficient design. In contrast, clients in FedAvg, FedNova, FedProx, and FedOpt consume 4.91, 3.99, 5.37, and 4.82 times more energy, respectively. The high energy efficiency of FedHDPrivacy stems from its computationally lightweight HD model, which reduces the overhead typically associated with transformer-based architectures.}

\textcolor{black}{Similarly, Figure~\ref{fig:bench_clientenergy2} presents the energy consumption comparison for CNN-based FL frameworks. While CNN-based models are generally more energy-efficient than transformer-based ones, FedHDPrivacy continues to demonstrate superior energy efficiency. Clients in FedAvg, FedNova, FedProx, and FedOpt consume 1.79, 1.82, 2.83, and 1.79 times more energy, respectively, compared to FedHDPrivacy. This consistent performance advantage highlights the framework's suitability for deployment in resource-constrained environments.}

\textcolor{black}{The results across all evaluation metrics—accuracy, training time, and energy efficiency—clearly demonstrate the superiority of FedHDPrivacy over traditional FL frameworks. By effectively balancing privacy, computational efficiency, and model performance, FedHDPrivacy provides a robust solution for FL in dynamic IoT environments.}

\section{Conclusions and Future Work} 
\label{sec:Conclusions and Future Work}
This paper presented the FedHDPrivacy framework, an innovative approach to addressing privacy concerns in IoT systems using a combination of FL, HD, and DP. The FedHDPrivacy framework ensures the protection of sensitive data while maintaining strong model performance in a distributed learning environment. By integrating DP mechanisms within the HD model, the framework enables precise control of the noise introduced at both the client and server levels, effectively balancing privacy and accuracy. One of its key strengths is its explainability, which allows for the careful calculation of required noise in each communication round, tracking cumulative noise from previous rounds and minimizing unnecessary noise in future iterations. This approach ensures that privacy is maintained without significantly degrading the model’s performance.

The FedHDPrivacy framework also addresses the challenge of continuous learning in IoT systems, ensuring that models remain up-to-date and adaptive in dynamic environments. FedHDPrivacy controls the added noise to the model, maintaining a balance between privacy and accuracy. This control is especially crucial in a lifelong learning context, where cumulative noise over time can degrade performance. This is particularly valuable in real-world applications, where IoT devices continuously generate data and require models that can evolve over time. Moreover, FedHDPrivacy provides a robust defense against common privacy threats such as model inversion and membership inference attacks, making it a practical solution for safeguarding sensitive information in FL setups.

For future work, exploring additional security measures, such as defenses against model posing and free-riding attacks, could further enhance the reliability of the system. Addressing these issues would help prevent dishonest participation in the FL process and improve the accuracy of the final model. Additionally, as IoT devices often record diverse data types, such as images, sensor signals, and textual data, expanding the FedHDPrivacy framework to handle multimodal data fusion—combining different data types for improved model performance—could offer significant advancements. Another promising avenue for future exploration would be the application of FedHDPrivacy in more complex IoT environments, such as edge computing, where low-latency and real-time processing are critical. \textcolor{black}{Finally, enhancing the framework to adapt to different privacy budgets dynamically, based on varying data sensitivity across clients, could provide more tailored privacy solutions.}

\section*{Acknowledgment}
 This work was supported by the National Science Foundation [grant numbers 2127780, 2312517]; the Semiconductor Research Corporation (SRC); the Office of Naval Research [grant numbers N00014-21-1-2225, N00014-22-1-2067]; the Air Force Office of Scientific Research [grant number FA9550-22-1-0253]; UConn Startup Funding, and generous gifts from Xilinx and Cisco. The authors gratefully acknowledged the valuable contributions from the Connecticut Center for Advanced Technology, in particular, Nasir Mannan, for sharing data for this research.

\section*{Supplementary Materials}
\textcolor{black}{The code used in this study is openly available for reproducibility and can be accessed at the following link: \href{https://github.com/FardinJalilPiran/FedHDPrivacy}{https://github.com/FardinJalilPiran/FedHDPrivacy}}.

\bibliographystyle{unsrt}  
\bibliography{references}

\end{document}